\documentclass[lettersize,journal]{IEEEtran} 
\usepackage{booktabs}
\usepackage[table,dvipsnames]{xcolor}
\usepackage{makecell}         
\usepackage{threeparttable}   
\definecolor{smgpurple}{RGB}{128,0,128} 

\usepackage{soul}
\usepackage{xcolor}
\definecolor{paperpurple}{HTML}{7A3ADB} 
\newcolumntype{Y}{>{\centering\arraybackslash}X} 
\usepackage{booktabs,makecell} 
\usepackage[dvipsnames]{xcolor} 
\usepackage{multirow} 
\usepackage{booktabs,tabularx,makecell} 
\newcolumntype{Y}{>{\centering\arraybackslash}X} 
\usepackage{hyperref} 
\renewcommand{\arraystretch}{1.08} 
\usepackage{amsmath,amsfonts} 
\usepackage{algorithmic} 
\usepackage{algorithm} 
\usepackage{array} 
\usepackage[caption=false,font=normalsize,labelfont=sf,textfont=sf]{subfig} 
\usepackage{textcomp} 
\usepackage{stfloats} 
\usepackage{url} 
\usepackage{verbatim} 
\usepackage{graphicx}
\usepackage{cite} 
\begin{document}

\title{SMGeo: Cross-View Object Geo-Localization with Grid-Level Mixture-of-Experts}

\author{Fan Zhang,\textit{ Senior Member}\textit{, IEEE}, Haoyuan Ren, Fei Ma, \textit{Member, IEEE}, Qiang Yin, \textit{Senior Member, IEEE},
and Yongsheng Zhou, \textit{Member, IEEE}~\IEEEmembership{}
\thanks{This work was supported in part by the National Natural Science Foundation of China under Grant No.62201027 and
 No.62271034. (Corresponding author: Fei Ma.) }
\thanks{
Fan Zhang is with the School of Information Science and Technology and the Interdisciplinary Research Center for Artificial Intelligence, Beijing University of Chemical Technology, Beijing 100029, China (e-mail: zhangf@mail.buct.edu.cn).}
\thanks{Haoyuan Ren, Fei Ma, Qiang Yin, and Yongsheng Zhou are with the School of Information Science and Technology, Beijing University of Chemical Technology, Beijing 100029, China (e-mail: 2023200784@buct.edu.cn; mafei@mail.buct.edu.cn; yinq@mail.buct.edu.cn; zhyosh@buct.edu.cn.)}}


\IEEEpubid{}

\maketitle
\begin{abstract}
Cross-view object geo-localization aims to precisely pinpoint the same object across large-scale satellite imagery based on drone images. Due to significant differences in viewpoint and scale, coupled with complex background interference, traditional multi-stage “retrieval-matching” pipelines are prone to cumulative errors. To address this, we present SMGeo, a promptable end-to-end transformer-based model for object geo-localization. This model supports click prompting and can output object geo-localization in realtime when prompted to allow for interactive use. The model employs a fully transformer-based architecture, utilizing a Swin Transformer for joint feature encoding of both drone and satellite imagery and an anchor-free transformer detection head for coordinate regression.
In order to to better capture both inter- and intra-view dependencies, we introduce a grid-level sparse Mixture-of-Experts (GMoE) into the cross-view encoder, allowing it to adaptively activate specialized experts according to the content, scale and source of each grid.  We also employ an anchor-free detection head for coordinate regression, directly predicting object locations via heatmap supervision in the reference images. This approach avoids scale bias and matching complexity introduced by predefined anchor boxes. On the drone-to-satellite task, SMGeo achieves leading performance in accuracy at IoU=0.25 and mIoU metrics (e.g., 87.51\%, 62.50\%, and 61.45\% in the test set, respectively), significantly outperforming representative methods such as DetGeo (61.97\%, 57.66\%, and 54.05\%, respectively). Ablation studies demonstrate complementary gains from  shared encoding, query-guided fusion, and grid-level sparse MoE.  (The code is available at https://github.com/KELE-LL/SMGeo)
\end{abstract}


\begin{figure}[t]
\centering
\includegraphics[width=\linewidth]{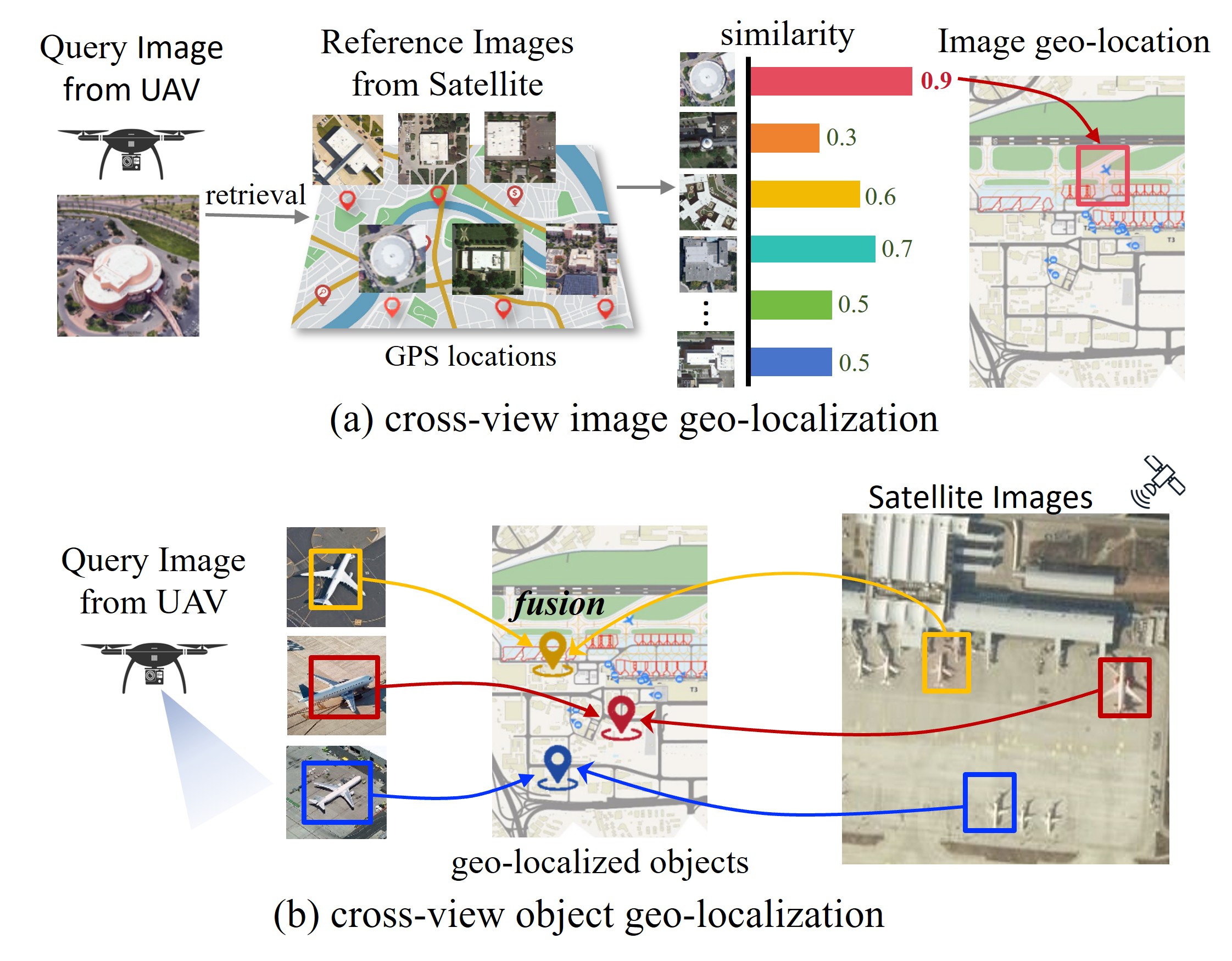}
\caption{Cross-view image geo-localization. (a) Traditional image-level matching methods. Given a query image, this approach retrieves the similar images from a large-scale satellite image database and returns a ranked list. (b) Object-level geo-localization. The task aims to geo-localize a specific object by using images captured from different viewpoints.}
\label{fig_1}
\end{figure}

\section{Introduction}
\IEEEPARstart{V}{ISUAL} geo-localization is a fundamental problem in computer vision: given an input image, the goal is to determine its geographic coordinates. It supports applications in autonomous driving vehicles, robotics and unmanned-system navigation \cite{ref1}. Visual geo-localization can be further divided into single-view and cross-view. The objective of single-view geo-localization is to estimate the GPS coordinates of a query image without using additional viewpoints. This is typically achieved by comparing the query image against a database of geotagged reference images. In contrast, cross-view geo-localization focuses on matching and localizing image pairs captured from different altitudes and viewpoints—e.g., low-altitude unmanned aerial vehicle (drone) views versus high-altitude satellite views—and has attracted growing attention for its practical value \cite{ref1,ref2}. In this setting, the query image is typically acquired by a low-altitude drone or a ground device, whereas the reference is drawn from wide-area remote-sensing imagery (e.g., satellite). The two domains exhibit pronounced discrepancies, including scale variation, viewpoint rotation, limited field-of-view overlap, occlusion, and inconsistencies in illumination and texture \cite{ref3}. Achieving reliable object correspondence and localization under such extreme cross-view differences remains highly challenging \cite{ref4,ref5,ref6}.

\graphicspath{{figs/}}  
\begin{figure}[!t]
\centering
\includegraphics[width=\linewidth]{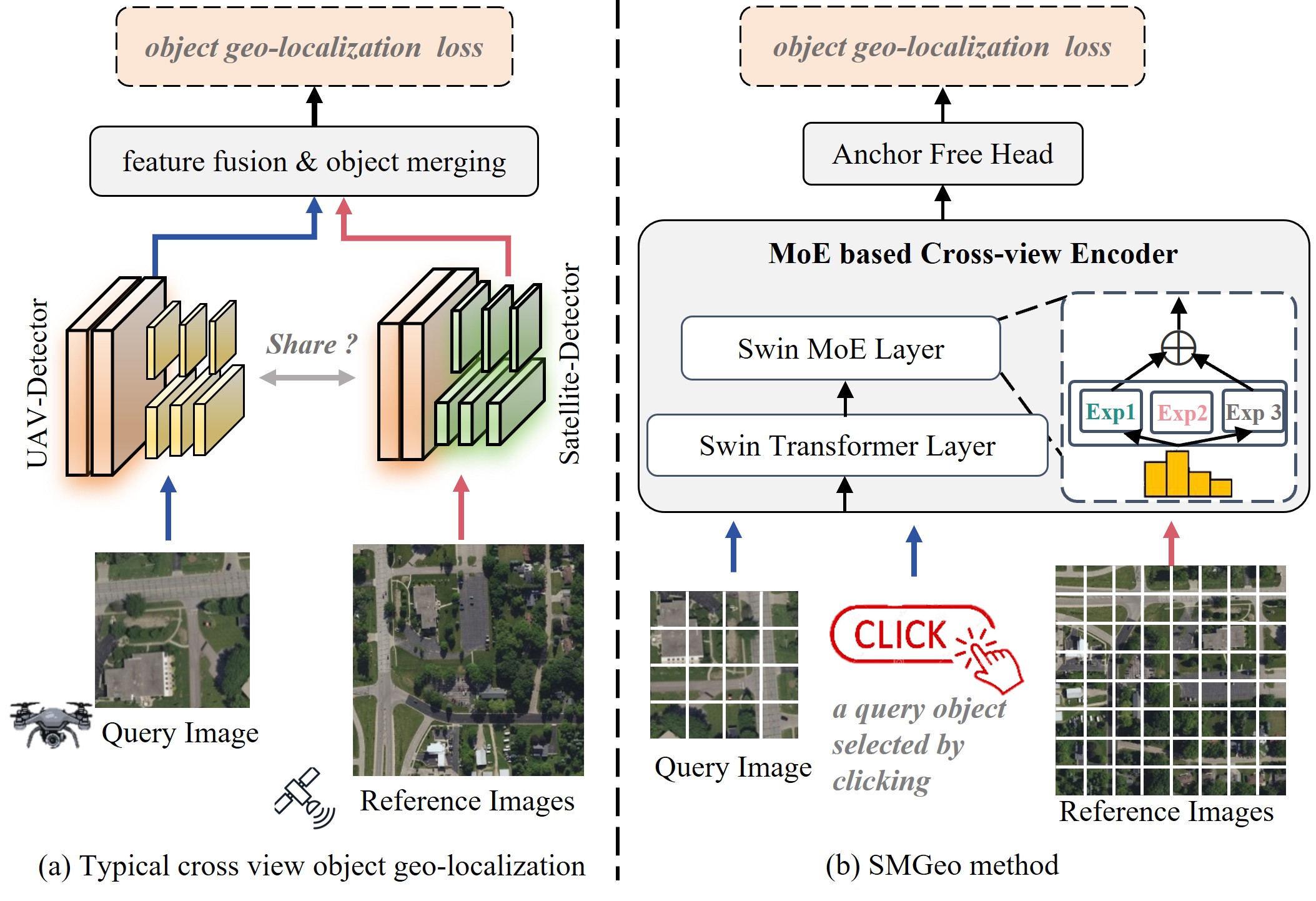}  
\caption{Structure comparison between the previous cross view object geo-localization and the proposed SMGeo. (a) Typical cross view object geo-localization. Given a certain view input, the latent representations are extracted by the specific encoder, followed by object detection and merging. (b) Our SMGeo method. Designed for interactive use, the model supports click prompting and delivers real-time localization results. SMGeo introduces a grid-level Mixture-of-Experts (GMoE) based cross-view encoder that jointly learns cross-view representations. The GMoE adaptively activates specialized experts to capture both inter- and intra-view dependencies. In addition, an anchor-free head directly regresses the target’s coordinates in the reference images.}
\label{fig_2}
\end{figure}

The current cross-view geo-localization approaches are primarily designed to predict the GPS coordinates of the entire imagery \cite{ref7,ref8,ref9,ref10}, as illustrated in Fig.~\ref{fig_1}(a). These approaches encode an entire query image into a global descriptor and perform similarity search within geotagged reference collections (typically satellite imagery). The retrieved reference coordinates are then used as the estimated location. To alleviate viewpoint discrepancies and semantic gaps, representative works have evolved from early handcrafted features and metric learning to deep global representations with Transformer-based cross-modal alignment (e.g., CVM-Net, GeoDTR, TransGeo). Such methods have achieved improvements in Top-k accuracy and generalization. Most employ dual-branch architectures, such as SIFT-based pipelines \cite{ref11,ref12}, LPN \cite{ref13}, GeoDTR \cite{ref14}, and TransGeo \cite{ref15}.  It has driven increasing interest in object geo-localization in recent years, which locates distinctive objects (like buildings or landmarks ) within scenes \cite{ref13,ref16}. The object geo-localization methods, as illustrated in Fig.~\ref{fig_1}(b), predict the GPS position of individual objects visible within the image. A fundamental challenge in object geo-localization is resolving duplicate detections of the same object across multiple frames. This necessitates mechanisms for deduplication, multi-view fusion, and deriving a single geographic coordinate. Mainstream approaches comprise three categories: tracker-based methods, triangulation-based methods, and re-identification or joint multi-view detection.

Tracker-based methods first generate candidate targets frame-by-frame using detectors, then utilize cross-frame correlation or multi-object tracking to group detections of the same object across frames into tracklets. Finally, each tracklet undergoes regularized fusion (e.g., weighted averaging) to derive single-point geographic coordinates \cite{ref17}. Chaabane \cite{ref18} proposed a “three-stage” near-end-to-end system. Stage 1 simultaneously detects objects and regresses their 5D poses (three translations + two rotations), then combines camera GPS data to obtain target coordinates via coordinate transformation. Stage 2 calculates cross-frame similarity matrices using geometric and appearance features. Stage 3 employs the Hungarian algorithm for optimal matching to form tracks. This approach offers high interpretability in controlled scenarios but heavily relies on precise target poses and camera intrinsic parameters—information often unavailable in remote sensing and drone platforms. Wilson et al. \cite{ref19} augmented RetinaNet with a GPS regression subnetwork to directly predict coordinates. Additionally, they train a similarity network to evaluate matching scores between adjacent frame target pairs, employing a thresholded Hungarian algorithm for cross-frame association. Coordinates are ultimately output as a weighted average for each trajectory. However, in cross-view scenarios involving drones and satellites, data typically appears in single-frame form, lacking continuous frames and stable camera parameters, rendering such methods impractical. Triangulation-based methods form triangles using two frames (two camera positions) and the angle between the camera and the target, estimating target distance via trigonometric functions and propagating it recursively through the sequence. To address noise, triangulation is often formulated as a least-squares or energy minimization problem for robustness. Krylov \cite{ref20} proposed a method that jointly optimizes CNN outputs for segmentation and depth estimation with MRF triangulation, combined with hierarchical clustering for redundancy elimination, balancing geometric interpretability and depth features. However, such methods generally underperform compared to deep learning-based direct coordinate regression approaches. More critically, triangulation heavily relies on multi-view inputs and stable geometric angles. Given the significant scale differences and extreme viewpoint variations between drone and satellite imagery, geometric constraints often fail to hold in practice. 

Joint multi-view detection methods aim to bypass cumbersome “tracking + fusion” processes. They simultaneously input multiple frames during detection, directly outputting single geographic coordinate predictions by integrating detection and re-identification, thereby implicitly merging redundant detections. Nassar \cite{ref21} employed a twin network to jointly learn detection, re-identification, and geo-localization for paired images. It first regresses bounding boxes on one frame, then learns cross-view projections to the other frame's perspective for registration and geo-localization, eliminating the need for additional trackers. This paper tackles object geo-localization from drone-satellite image pairs. In this scenario, tracker-based methods are unsuitable due to the lack of sequential frames, and triangulation-based methods fail because of the significant altitude difference between drones and satellites \cite{ref22}. Given these constraints, joint multi-view detection methods present a more suitable approach for object localization in drone-satellite image pairs. However, our analysis of existing joint cross-view detection methods reveals two primary limitations \cite{ref23}.

First, most joint multi-view detection methods rely on CNNs for feature representation and fusion, which restricts their robustness and generalization in complex environments (Fig. \ref{fig_2}(a)). In contrast, the transformer architecture better aligns with the task requirements for global feature modeling and cross-domain alignment. Specifically, on one hand, the self-attention mechanism enables global receptive fields that adapt to content, allowing distant objects and context to be correlated within a single frame, thereby mitigating the non-local matching challenges posed by large parallax. On the other hand, cross-attention mechanisms facilitate explicit feature interaction and alignment between drone and satellite images during the encoding phase, avoiding representation shifts caused by post-fusion of dual streams \cite{ref24}. Furthermore, hierarchical transformers (e.g., Swin Transformer) employ windowed self-attention to preserve multi-scale details while effectively controlling computational complexity, enabling complementary modeling of local information and global relationships in high-resolution images. The introduction of relative position encoding provides learnable geometric priors, further enhancing robustness to viewpoint and scale variations. In summary, the advantages of transformers in global modeling and cross-domain interactions precisely meet the core requirements of robust feature representation and coordinate prediction for drone-satellite single-frame cross-view object localization. 

Secondly, joint multi-view detection models typically employ a dual-branch architecture to separately extract features of query and reference images. The dual-path CNN framework proposed by Vo et al. ~\cite{ref25} is a representative example of this approach. Due to the lack of parameter sharing and coordination between the two branches, dual-branch architectures often suffer from inconsistent feature representation spaces, redundant model parameters, and high deployment complexity. To address these limitations, some studies have begun exploring unified modeling approaches using a “shared backbone network”. For instance, DINO-MSRA ~\cite{ref26} introduces a matching attention mechanism on a shared visual encoder to achieve whole-image feature alignment. Huang et al. ~\cite{ref27} combine visual-language pre-training strategies to capture cross-domain semantic consistency. However, most current methods still follow dual-branch or whole-image matching paradigms, and end-to-end unified architectures directly targeting “object-level geo-localization” remain rare.

\begin{figure}[!t]  
\centering
\includegraphics[width=\linewidth]{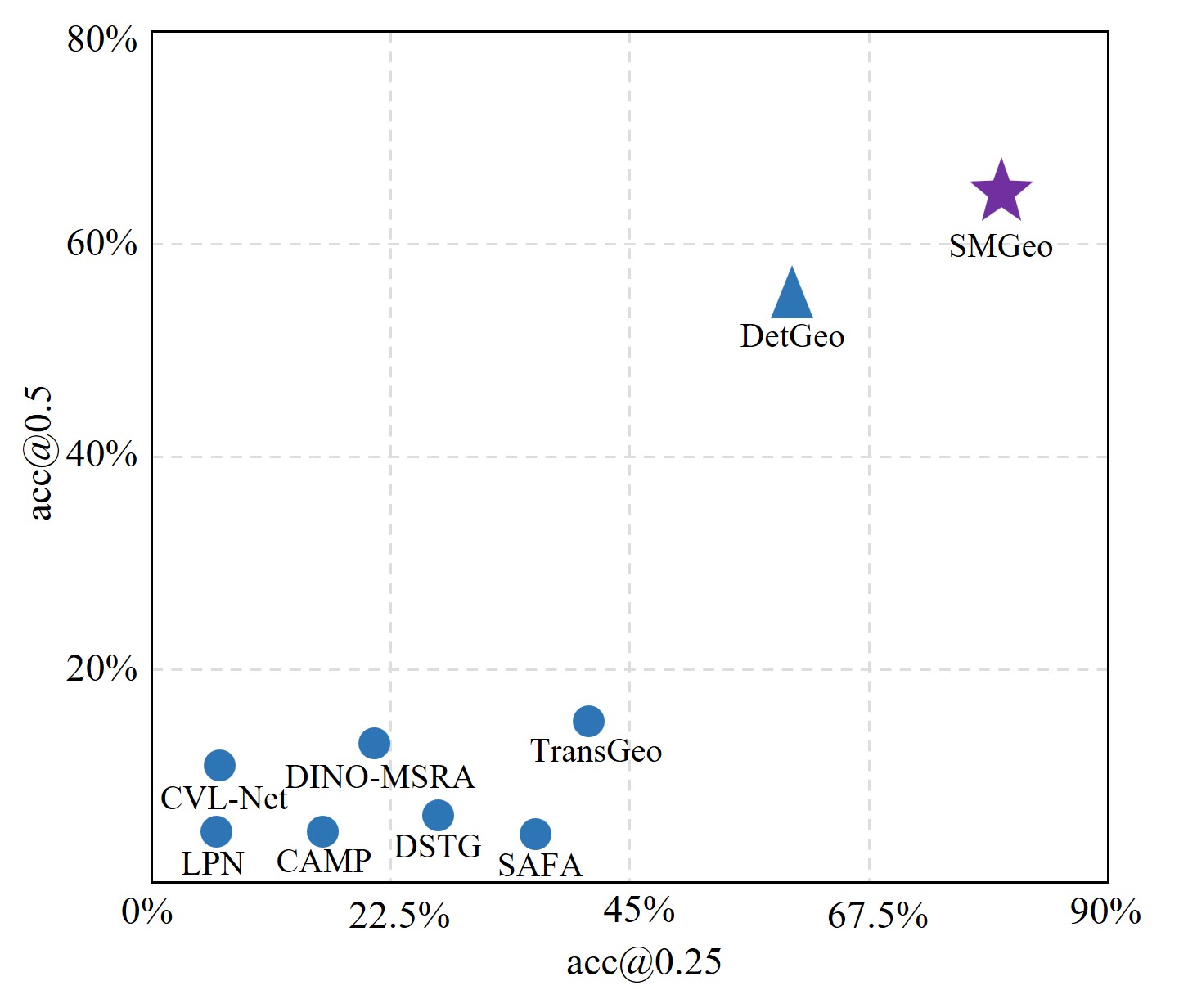}
\caption{Comparison of geo-localization accuracy among different cross-view methods on the CVOGL Dataset. The horizontal axis represents the acc@0.25, and the vertical axis represents the acc@0.5. Our SMGeo method significantly outperforms existing methods on both metrics, demonstrating optimal object geo-localization performance.}
\label{fig_3}
\end{figure}

Based on the above insights, this paper proposes a transformer-based cross-view object geo-localization with grid-level Sparse Mixture-of-experts (SMGeo). As shown in Fig.~\ref{fig_2}(b), this method employs a swin transformer with shared weights as its backbone, simultaneously performing feature encoding and semantic alignment on both the query image from the drone perspective and the reference image from the satellite perspective. Leveraging mechanisms such as windowed self-attention, the model effectively integrates local details with global context, \cite{ref28,ref29}. Simultaneously, addressing the insufficient expressive rigidity of traditional feed-forward network (FFN) in cross-view tasks, this paper replaces portions of the FFN with a grid-level sparse mixture-of-experts (GMoE) model, enhancing the network's dynamic adaptability to different regional semantics. Previous studies (e.g., SM3Det \cite{ref30,ref31}) have also validated the effectiveness of a grid-level spare MoE in multimodal object detection, further substantiating the rationality of this structural design. To optimize spare MoE's routing stability and expert distribution diversity, we introduce a gated distribution entropy regularization term into the loss function. This enhances routing discriminativeness while preventing excessive concentration of expert paths \cite{ref32}. We conduct a systematic evaluation on the challenging real-world cross-view object localization dataset (CVOGL\_DroneAerial) \cite{ref33}. Experimental results, shown in Fig.~\ref{fig_3}, demonstrate that our method significantly outperforms existing mainstream approaches across multiple localization accuracy metrics.

The contributions of this paper can be summarized as follows:  
\begin{enumerate}
    \item We present SMGeo, a promptable end-to-end transformer-based model for object geo-localization. This model supports click prompting and can output object geo-localization in realtime when prompted to allow for interactive use. The model employs a fully transformer-based architecture, utilizing a Swin Transformer for joint feature encoding of both drone and satellite imagery and an anchor-free transformer detection head for coordinate regression.
    
    \item We propose a grid-level sparse Mixture-of-Experts (GMoE) cross-view encoder that jointly learns cross-view representations. The input cross-view images or features are divided into multiple grids, and a grid-level routing module is introduced into the encoder backbone. This enables the encoder to adaptively activate specialized experts according to the content, scale and source of each grid, allowing the model to better capture both inter- and intra-view dependencies.
    
    \item We employ an anchor-free detection head for coordinate regression, directly predicting object locations via heatmap supervision in the reference images. This approach avoids scale bias and matching complexity introduced by predefined anchor boxes, enhancing localization accuracy and generalization while maintaining structural simplicity.
    
\end{enumerate}

The remainder of this paper is organized as follows. Section~\ref{sec:related} reviews prior work on cross-perspective geo-localization. Section~\ref{sec:methodology} details our methodology. Section~\ref{sec:experiments} presents our comprehensive experimental results and analysis. Finally, we provide a summary of this paper in Section~\ref{sec:conclusion}.

\section{Related Works}\label{sec:related}
\subsection{Cross-View Geo-localization}

The majority of current cross-view geo-localization research remains focused on image-level matching, which directly predicts the location of the query image within a large-scale reference image. Typical approaches such as CVM-Net \cite{ref16}, LPN \cite{ref13}, and TransGeo \cite{ref15} all build matching frameworks based on whole-image feature similarity, suitable for coarse-grained localization or retrieval tasks.

However, in practical application scenarios such as target monitoring, navigation guidance, and disaster response, more refined “object-level” geo-localization capabilities are often required. Such approaches typically demand the construction of finer-grained feature representations to accurately perceive and match specific geographic entities (e.g., roads, buildings, bodies of water) \cite{ref34}. To address this, some research has begun evolving from whole-image modeling toward object-level representations. For instance, Zhu et al. \cite{ref2} enhanced cross-view region alignment by constructing a semantically consistent embedding space; TransGeo \cite{ref15} introduced context-aware mechanisms to capture structural information, enabling more robust feature matching in local regions; Toker et al. \cite{ref35} reconstructed and associated cross-view semantic domains using polar coordinate transformations and image generation models.

Although the aforementioned methods mitigate representation bias caused by viewpoint differences to some extent, most still rely on whole-image matching frameworks, failing to fully leverage object-level structure information within images. This limitation significantly degrades localization accuracy, particularly when occlusions, scale variations, or semantic ambiguity are present. Recent studies have introduced novel approaches. For example, Guo et al. \cite{ref36} enhanced region correspondence through geometric consistency modeling; relying on reliable local semantic segmentation and stable geometric priors. However, in scenarios like drone–satellite imaging—characterized by single frames, drastic scale and viewpoint variations—these assumptions often fail. The SMGeo model proposed in this paper incorporates a grid-level sparse MoE that can adaptively learn spatial topological relationships across multi-view images.

In summary, while object-level cross-view localization is gaining attention, current research still faces limitations in regional modeling accuracy and object-level positioning capability. The proposed SMGeo model addresses this gap, aiming to achieve precise localization of key target regions while maintaining global perception.

\subsection{Mixture-of-Experts (MoE)}

The multi-expert hybrid mechanism is a strategy that enhances the representational capacity and generalization performance of neural networks by enabling multiple subnetworks (i.e., experts) to collaboratively participate in computations. The soft-gated MoE architecture, initially proposed by Jacobs et al. ~\cite{ref37,ref38}, achieves dynamic path selection by training a gating network to assign activation weights to each expert based on input features. Shazeer et al. ~\cite{ref39} introduced sparse gating to activate only the top-k experts, significantly reducing computational overhead and improving training efficiency, making MoE a key technology for scaling large-scale models. In computer vision, MoE applications have expanded to image classification, object detection, and multimodal learning tasks. Riquelme et al. ~\cite{ref40} pioneered integrating MoE into convolutional neural networks, validating the expressive advantages of the expert architecture in complex image scenarios. Under the transformer framework, MoE further employs a token-by-token dynamic allocation mechanism, enabling optimal expert assignment for feature tokens across different spatial locations or semantic categories. This achieves personalized feature representation and fine-grained feature modeling.

For remote sensing imagery and multi-view scenarios, MoE inherently offers distinct advantages. On one hand, images captured from drone and satellite perspectives exhibit significant differences in semantic content and structural distribution, making unified models prone to confusing features from different viewpoints. On the other hand, the uneven distribution of semantic information across regions demands models with dynamic perception and feature reconstruction capabilities. To address this, we introduce a grid-level spare MoE mechanism within the swin transformer backbone. Each FFN sublayer is replaced with a module composed of multiple experts. Combined with a top-k expert selection strategy based on gating probabilities and an entropy regularization term, this enables the model to activate the most suitable subnetwork for different image regions. This effectively enhances local feature perception and cross-view robustness. Inspired by the SM3Det ~\cite{ref30} approach, our work extends the spare MoE mechanism for the first time to cross-view object-level localization under a unified backbone network, and proposes a grid-level expert activation strategy to enhance the model's dynamic adaptability to regional feature variations ~\cite{ref31}. 

\graphicspath{{figs/}}
\begin{figure*}[!t]
\centering
\includegraphics[width=\textwidth]{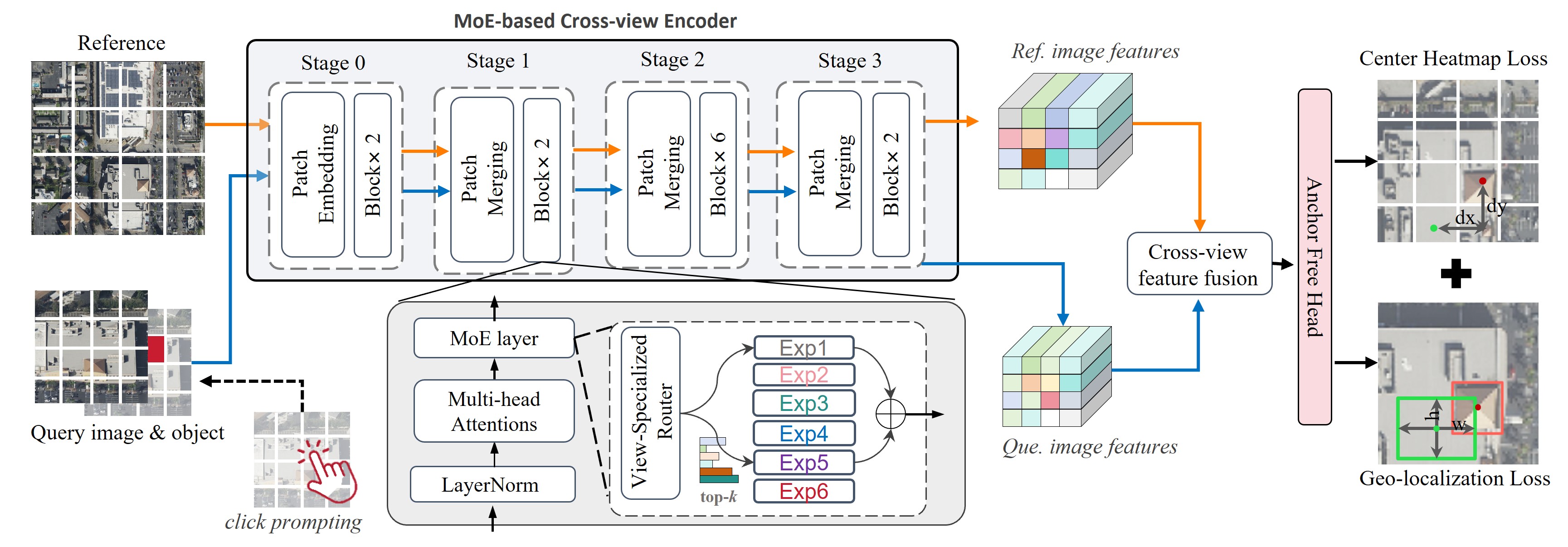}  
\caption{Overall framework of the proposed SMGeo. A cross-view encoder based on GMoE utilizes a view-specific router for top-k expert selection to adaptively process cross-view image features. Subsequently, a cross-view feature fusion module fuses the features from the two views with the encoded click prompts. Finally, the anchor-free detection head directly regresses target center heatmaps and bounding box offsets.}
\label{fig_4}
\end{figure*}

\section{Methodology}\label{sec:methodology}

\subsection{Overview}

As shown in Fig.~\ref{fig_4}, the overall architecture of SMGeo consists of three components: a GMoE-based cross-view encoder, a dynamic feature fusion module guided by the semantic information from GMoE, and a cross-view localization head. First, we propose a GMoE based cross-view encoder that can adaptively activate specialized processing experts for multi-view inputs within a unified backbone network. Concurrently, we formulate an adaptive fusion strategy that adaptively performs semantic selection and fusion based on image content, thereby addressing the limitations of existing approaches in terms of flexibility for cross-domain regional representation and feature selection capabilities. Finally, by designing an anchor-free detection head, we further simplify the model structure, eliminating the interference of traditional anchor-based matching strategies on cross-view scale perception. This provides an efficient and feasible solution for cross-view object geo-localization.

\subsection{GMoE-Based Cross-View Shared Encoder}
To achieve semantic alignment and information sharing across cross-view image pair, we employ an enhanced swin transformer as the shared backbone network for unified feature extraction and modeling of drone-satellite image pairs. Compared to CNN \cite{ref41}, the swin transformer employs a shifted window self-attention mechanism. This approach effectively captures long-range dependencies while preserving local details, combining local modeling capabilities with global information awareness. Leveraging this property, the swin backbone achieves structural alignment and semantic abstraction across images from different viewpoints, avoiding the issue of feature semantic inconsistency found in dual-branch networks.

Given that this study focuses on target localization in drone and satellite imagery, we treat the input as a cross-view image pair. To reduce unnecessary redundancy and enhance efficiency, We designed a unified cross-view encoder. As shown in Fig.~\ref{fig_4}, both the query image and reference image are fed into a shared swin transformer backbone for feature encoding. This structural approach avoids the parameter redundancy and heterogeneous representation issues inherent in dual-branch architectures. This strategy not only significantly simplifies the model structure but also improves computational efficiency while ensuring feature consistency \cite{ref42,ref43}.

\begin{figure}[t]
\centering
\includegraphics[width=\linewidth]{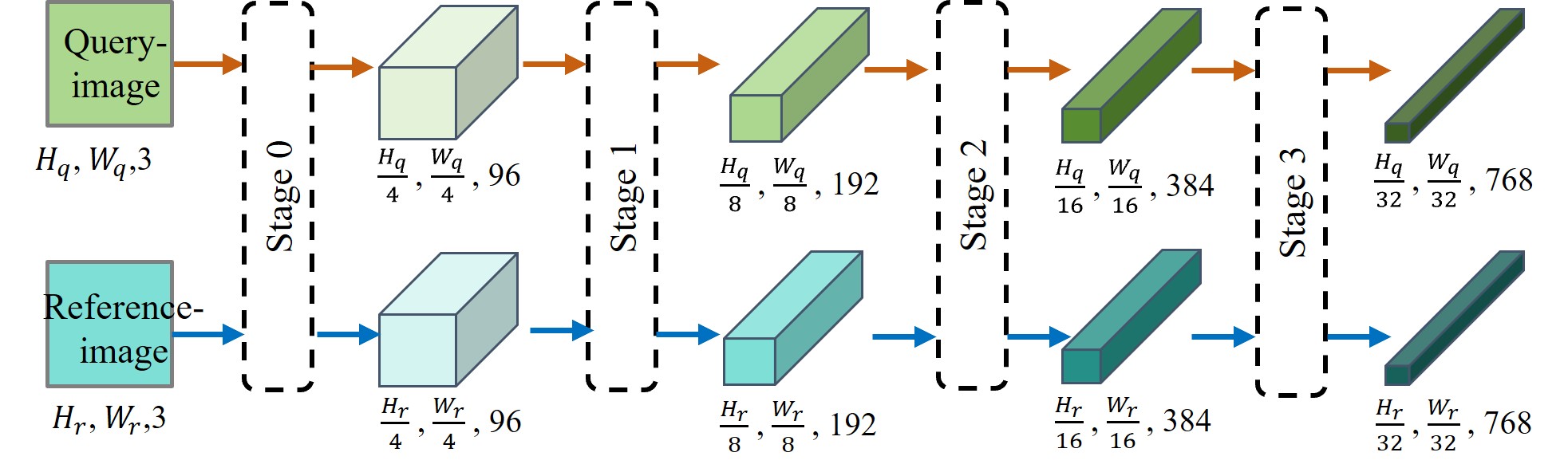}
\caption{Tracking the size changes of the query image and reference images.}
\label{fig_5}
\end{figure}

For the input query image and reference image, the original image is segmented into multiple $P \times P$ patches through a convolution operation, thereby converting it into a sequence of token features. As an example, let the input reference image be $I_r \in \mathbb{R}^{H_r \times W_r \times C}$. Through the patch embedding layer, this can be defined as:
\begin{equation}
I_r^{PE} = \text{Conv}(x_r, P, P), \quad I_r^{PE} \in \mathbb{R}^{\left( \frac{H_r W_r}{P^2} \right) d}
\end{equation}
where $P$ denotes the patch size and $d$ represents the embedding dimension. Subsequently, the token features undergo normalization via LayerNorm to yield the initial token sequence:
\begin{equation}
X_r^{token} = \text{LayerNorm}(I_r^{PE})
\end{equation}

Subsequently, $X_r^{token}$ enters the swin transformer backbone network, which consists of four stages. Each block consists of two parts: window-based multi-head self-attention (W-MSA) or shifted window-based multi-head self-attention (SW-MSA) and a multilayer perceptron (MLP). W-MSA models self-attention within local windows, while SW-MSA enhances cross-window information exchange. For the token sequence $X_r^{\text{token}} \in \mathbb{R}^{\left( \frac{H_r W_r}{P^2} \right) d}$ (where $\frac{(H_r \times W_r)}{P^2}$ is the number of tokens), the attention operation at layer $l$ is expressed as:
\begin{equation}
\text{Attention}_l(Q_l, K_l, V_l) = \text{Softmax}\left( \frac{K_l^\top \cdot Q_l}{\sqrt{d}} + B_l \right) \cdot V_l
\end{equation}
where $Q_l = X_{r,l-1}^{\text{token}} \cdot W_{\text{Q}}^l \quad$, $K_l = X_{r,l-1}^{\text{token}} \cdot W_{\text{K}}^l \quad$, $V_l = X_{r,l-1}^{\text{token}} \cdot W_{\text{V}}^l$. $W_{\text{Q}}^l$, $W_{\text{K}}^l$ and $W_{\text{V}}^l$ are the linear transformation weight matrix for the $l$-th layer block. $X_{r, l-1}^{\text {token }}$ is the input to the $l$-th layer block, with output $X_{r, l}^{\text {token }}$                 denote query, key, and value matrices, $B_l$ is the relative positional bias, and $d$ is the dimensionality per head.
This design achieves efficient modeling of cross-spatial relationships while ensuring local perceptual capabilities. In contrast, the window attention mechanism within the swin transformer restricts attention computations to a local window. Taking a window size of $M$ × $M$ as an example, for the input $X_{r, l-1}^{\text {token }}$ of the $l$-th layer block, the window attention mechanism can be expressed as: 
\begin{equation}
\begin{split}
X_{r,l-1}^{\text{token},\text{W-MSA}}
&= \text{W-MSA}\big(\text{LayerNorm}(X_{r,l-1}^{\text{token}})\big) \\
&\quad + X_{r,l-1}^{\text{token}}
\end{split}
\end{equation}
where $\operatorname{W\!-\!MSA}$ denotes the window-based multi-head self-attention. Subsequently, the feature transformation is carried out by the \text{MLP} module:
\begin{equation}
\begin{split}
X_{r,l-1}^{\text{token, MLP}}
&= \text{MLP}\big(\text{LayerNorm}\big(X_{r,l-1}^{\text{token, W-MSA}}\big)\big) \\
&\quad + X_{r,l-1}^{\text{token, W-MSA}}
\end{split}
\end{equation}
where the MLP is defined as:
\begin{equation}
\text{MLP}(X_{r,l-1}^{\text{token, W-MSA}}) = FC_2(\text{GELU}(FC_1(X_{r,l-1}^{\text{token, W-MSA}})))
\end{equation}
where $FC_1$ and $FC_2$ denote fully connected layers.

The above design efficiently models cross-view feature correlations while preserving local perception capabilities. The swin transformer backbone employs a patch merging module at the end of each stage to downsample feature maps, progressively reducing spatial dimensions while increasing channel dimensions. This process generates multi-scale, layer-wise semantic representations. As shown in Fig.~\ref{fig_5}, for example, an input image of size $1024 \times 1024$ undergoes patch embedding and four-stage backbone extraction, reducing the spatial dimensions of feature maps successively to $1/4$, $1/8$, $1/16$, and $1/32$ of the input size (corresponding to dimensions like $256 \times 256$, $128 \times 128$, $64 \times 64$, and $32 \times 32$), while the channel dimension increases accordingly (e.g., 96, 192, 384, 768), yielding robust, multi-scale representations. By performing feature extraction on both query and reference images through a shared unified backbone network, we achieve natural alignment and consistent modeling of the feature space. This approach avoids semantic shifts that can arise from using different network architectures or parameter initializations. The shared backbone encourages the model to learn cross-view feature representations, establishing a stable foundation for subsequent adaptive fusion and localization prediction.

\begin{figure}[t]
\centering
\includegraphics[width=\linewidth]{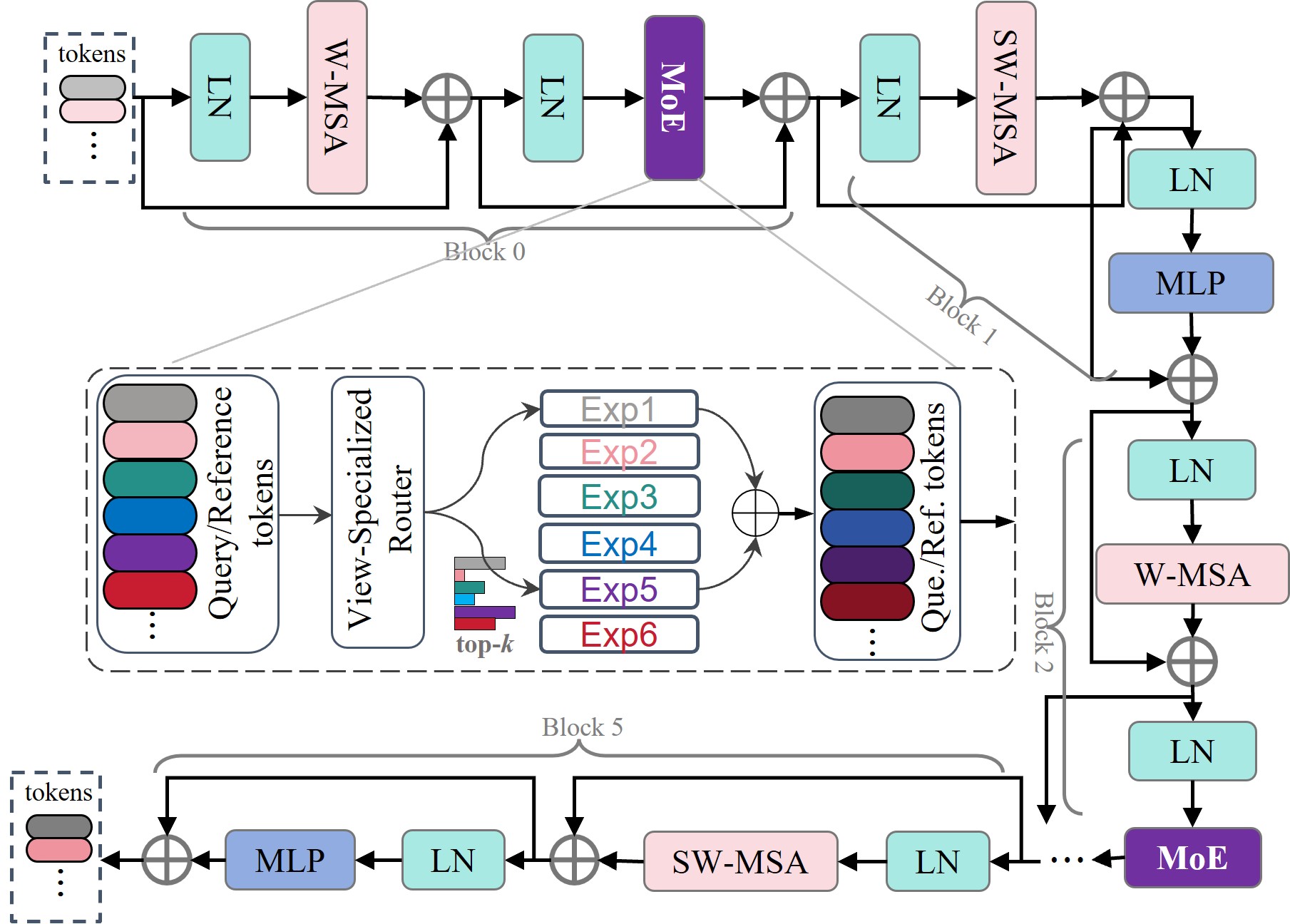}
\caption{The GMoE details in cross-view encoders. Taking the third stage as an example, we insert GMoE into blocks with even indices (i.e., inserting GMoE into Block 0, Block 2, and Block 4).}
\label{fig_6}
\end{figure}

\subsection{Adaptive Cross-View Feature Fusion Based on Grid-Level Mixture of Experts}
Although the shared backbone alleviates inconsistencies in feature space distribution between cross-view images, significant differences in imaging view, image size, and content complexity still lead to regional semantic misalignment errors and background noise interference between the query image and reference image \cite{ref44}. Traditional MLP relies solely on global features extracted through a single path, which struggles to capture all potential cross-domain variations. Simultaneously, conventional feature fusion strategies struggle to effectively align semantic information across dual perspectives, often leading to inconsistent regional perception and limiting the expressive power of target features. To this end, we introduce a “special” feature fusion mechanism. We employ the GMoE to replace the backbone MLP for region-adaptive encoding, combined with deep interactions between query and reference images to achieve region-adaptive feature modeling \cite{ref45,ref46}. As shown in Fig.~\ref{fig_6}, SSMGeo employs a combined MLP and GMoE in its middle layer to adaptively fuse cross-view image features.

Unlike traditional globally shared expert modules, our GMoE divides extracted 2D feature maps into a fixed number of spatial grids based on spatial location, with each grid corresponding to a local feature subregion. As an example of the input reference image, the output $X_{r}^{\text{token}} \in \mathbb{R}^{\left( \frac{H_r W_r}{P^2} \right) \times d}$ of the $l$-th layer block. We transform it into a two-dimensional feature map $X_{r,l} \in \mathbb{R}^{\frac{H_r}{P} \times \frac{W_r}{P} \times d}$, and subdivided into $G = g_h \times g_w$ grid cells, each containing characteristic sub-blocks of the local region. As an example, consider the feature $x_{r,l,g} \in \mathbb{R}^{{h_g} \times {w_g} \times d}$ of the $g$-th grid region (where $h_g = \frac{\dfrac{H_r}{P}}{g_h}, \quad w_g = \frac{\dfrac{W_r}{P}}{g_w}$). We designed multiple expert subnetworks $\left\{E_{i}(\cdot)\right\}_{i=1}^{k}$ and a gating routing network $g(x)$ to jointly achieve dynamic modeling and selective transformation of local feature regions. The output of the GMoE module is the weighted sum of outputs from each expert subnetwork:
\begin{equation}
\hat{y}_{\text{MoE}} = \sum_{i=1}^{k} g_i\left( x_{r,l,g} \right) \cdot E_i\left( x_{r,l,g} \right)
\end{equation}
where $E_i\left( x_{r,l,g} \right)$ is the processing output of the $i$-th expert network for the $l$-th layer block sub-block input features of the reference image, $g_i\left( x_{r,l,g} \right)$ is the corresponding gating weight dynamically generated by the gating network to control the contribution level of each expert to the output, and $k$ is the total number of experts. $E_i$ denotes the $i$-th activated expert network.The above mechanism enables each spatial region to select the most suitable expert for computation based on its own characteristic patterns, achieving dynamic modeling and selective enhancement of local areas.

To enhance the sparsity and stability of gate-controlled routing selection, we introduce a top-$k$ strategy that activates only the top $k'$ experts with the highest scores to participate in output computation, while resetting the output weights of the remaining experts to zero:
\begin{equation}
\begin{split}
g_t^{(\text{top}k)}(x_{r,l,g})
&= \frac{e^{g_i(x_{r,l,g})}}{\sum_{j \in \text{top}k\big(g(x_{r,l,g})\big)} e^{g_j(x_{r,l,g})}}, \\
&\qquad i \in \text{top}k\big(g(x_{r,l,g})\big).
\end{split}
\end{equation}
and the final output becomes:
\begin{equation}
\hat{y}_{\text{MoE}}^{\text{top}k} = \sum_{i \in \text{top}k\left(g(x_{r,l,g})\right)} g_i^{(\text{top}k)}(x_{r,l,g}) \cdot E_i(x_{r,l,g})
\end{equation}

After applying the top-$k$ strategy, the output of the GMoE module can be updated to contain only the weighted sum of the top-$k'$ expert contributions. This mechanism enhances the discriminative power and interpretability of expert selection, avoids redundant participation of irrelevant experts in computations, and improves the efficiency of feature fusion.   

\begin{figure}[t]
\centering
\includegraphics[width=\linewidth]{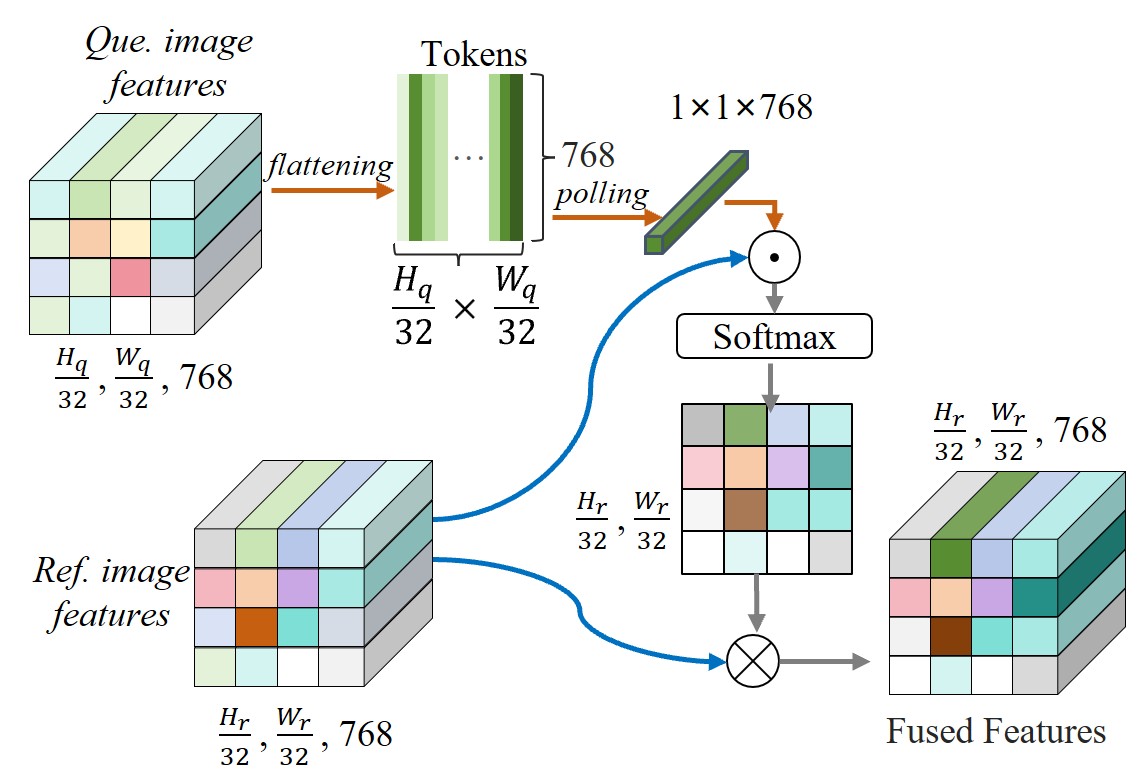}
\caption{Cross-view image feature fusion process. The query feature is flattened and undergoes global average pooling to produce a query vector. This vector is then compared with the reference feature to compute similarity scores, generating a spatial attention map. This map is used to weight the reference features, resulting in the final fused features.}
\label{fig_7}
\end{figure}

After obtaining multi-scale features of the query image and reference image in a unified space, we require further interactive fusion of these features to generate an aligned, object-aware fusion representation for downstream localization heads (as shown in Fig.~\ref{fig_7}). In simple terms, by using the features of the query image as guidance, adaptive weights are assigned to the local features of the reference image, enabling target-driven region matching and feature enhancement \cite{ref33}, \cite{ref47}. Let $F_q$ denote the global features extracted from the query image, and $F_r$ denote the spatial features extracted from the reference image. We first perform spatial pooling on $F_q$ to generate the global semantic vector $\bar{F}_q$. Then, through a dot-product mechanism, we compute weight coefficients by interacting with each spatial location in $F_r$:
\begin{equation}
A = \sigma\left(\bar{F}_q^\top \cdot F_r^{(i,j)}\right)
\end{equation}
where $A$ denotes the fusion attention map, and $\sigma(\cdot)$ represents the sigmoid activation function. Through this process, the query image can adaptively perceive the regional saliency distribution of the reference image, thereby guiding the spatial fusion process. 

It is worth emphasizing that, due to the shared backbone and content-aware routing strategy, the GMoE performs cross-processing of features from different views at the expert dimension when selecting local tokens at each layer. The same expert is likely to receive inputs from both the query and reference simultaneously, thereby establishing implicit cross-view associations within the functional space. In other words, GMoE's sparse routing and internal transformation process inherently constitute a continuous dynamic fusion mechanism. Through the coupling of expert selection and attention mechanisms, it progressively strengthens shared representations across views while reducing domain-specific biases. The explicitly designed fusion module then supplements these implicit interactions at the high-level semantic, significantly enhancing localization accuracy and robustness in cross-view geo-localization.

\subsection{Cross-view Localization Head Design}
After completing backbone feature extraction and feature fusion, we designed an efficient cross-view object localization detector to predict the precise location of objects in reference images. Unlike traditional anchor-based detection methods, our approach employs an anchor-free detector \cite{ref48,ref49} to directly regress target locations, eliminating the need for predefined anchor boxes and matching processes. This approach offers greater flexibility in adapting to scale uncertainties and appearance variations of cross-view objects. The anchor-free design is particularly well-suited for target localization tasks between drone and satellite imagery. The anchor-free localization head takes the cross-view fused spatial feature map as input and outputs two key results:
\begin{itemize}
    \item Center Heatmap: indicating the probability distribution of object centers across spatial locations;  
    \item Bounding Box Regression: for predicting the offset of the object center relative to the grid as well as the width and height of the object. 
\end{itemize}

Specifically, given the spatial feature map $F \in \mathbb{R}^{H' \times W' \times C'}$ after cross-view fusion, the anchor-free detection head consists of several convolutional layers that progressively analyze the features to extract the information required for localization:       
\begin{equation}
F_{feat} = \text{ReLU}(\text{BN}(\text{Conv}(\text{ReLU}(\text{BN}(\text{Conv}(F)))))))
\end{equation}
Based on $F_{feat}$, we produce the heatmap and bounding box predictions:
\begin{equation}
\text{Heatmap} = \sigma(\text{Conv}_{1\times1}(F_{feat})), \quad 
\text{BBox} = \text{Conv}_{1\times1}(F_{feat})
\end{equation}
where $\sigma(\cdot)$ is the sigmoid activation function. The center heatmap $(\mathrm{Heatmap} \in \mathbb{R}^{H \times W \times 1})$ represents the probability that each location belongs to the object center; the bounding box prediction $\mathrm{BBox}$ encodes the offset $(dx, dy)$ relative to the grid location of the center, as well as the object width and height $(w, h)$. 

To train the anchor-free detector, we need to construct supervision signals based on ground truth bounding boxes. Given the true bounding boxes of objects in an image:
\begin{equation}
\text{BBox}^{gt} = [x_1,y_1,x_2,y_2]
\end{equation}
First, map the bounding box of the real target onto the feature map scale and compute the center point:
\begin{equation}
c_x = \frac{x_1 + x_2}{2} \cdot \frac{W'}{W_r}, \quad
c_y = \frac{y_1 + y_2}{2} \cdot \frac{H'}{H_r}
\end{equation}
Then compute the Gaussian heatmap around the center point:
\begin{equation}
\text{Heatmap}_{gt}(x,y) = \exp\left(-\frac{(x-c_x)^2 + (y-c_y)^2}{2\sigma^2}\right)
\end{equation}
where $\sigma$ controls the spread of the positive region. The bounding box regression targets are defined as:
\begin{equation}
dx = c_x - \lfloor c_x \rfloor, \quad 
dy = c_y - \lfloor c_y \rfloor
\end{equation}
\begin{equation}
w = (x_2 - x_1)\frac{W'}{W_r}, \quad 
h = (y_2 - y_1)\frac{H'}{H_r}
\end{equation}
where $(c_x, c_y)$ denote the object center coordinates, $(x_1, y_1, x_2, y_2)$ are the bounding box corners, and $W_{img}, H_{img}$ represent the image width and height, respectively. $\lfloor \cdot \rfloor$ denotes the grid cell containing the center point.  

In summary, the anchor-free design philosophy not only effectively addresses the challenges posed by scale variations in cross-view targets but also provides greater flexibility and accuracy for end-to-end model training, significantly enhancing the performance of cross-view object localization. 

\subsection{Training Objective and Optimization Strategy}
To train SMGeo more effectively, this paper designs an optimization function comprising multiple sub-objectives within an end-to-end unified network architecture. This function encompasses cross-view feature representation, region candidate prediction, and gated routing \cite{ref45}, \cite{ref46}. The overall training process is based on a multi-task loss function, achieving efficient collaboration among different modules through joint optimization. The final total loss function consists of the following three components:

\begin{itemize}
    \item Focal Loss for center heatmap prediction ($L_{hm}$);
    \item L1 loss for bounding box regression ($L_{bbox}$);
    \item Entropy regularization for GMoE gating distribution ($L_{entropy}$).
\end{itemize}

The total loss is defined as:
\begin{equation}
{L}_{total} = \alpha \cdot {L}_{hm} 
+ \mu \cdot {L}_{bbox} 
+ \lambda \cdot {L}_{entropy},
\end{equation}
where $\alpha$, $\mu$, and $\lambda$ represent the weight coefficients for each sub-loss term, used to balance the importance of different training objectives. Their specific values will be detailed in the experimental section.

\paragraph{Center Heatmap Loss}  
To guide the anchor-free detection head in better predicting the center position of target regions, this paper employs the classic focal loss to construct a center point heatmap loss, effectively mitigating the imbalance between positive and negative samples. Its expression is as follows:
\begin{equation}
\begin{aligned}
{L}_{hm} ={} & -\frac{1}{N_{pos}} \sum_{x,y} 
\\ & 
\begin{cases}
(1-p_{xy})^{\gamma} \log(p_{xy}), & \text{if } y_{xy}=1, \\[6pt]
(1-y_{xy})^{\beta} (p_{xy})^{\gamma} \log(1-p_{xy}), & \text{otherwise}.
\end{cases}
\end{aligned}
\end{equation}
where $p_{xy}$ is the predicted heatmap value, $y_{xy} \in \{0,1\}$ indicates whether the position corresponds to a target center, $\gamma$ and $\beta$ are weighting parameters for hard/easy samples, and $N_{pos}$ is the number of positive samples.

\paragraph{Bounding Box Regression Loss}  
For regressing the bounding box size from candidate centers, we employ an L1 loss to minimize the difference between predicted and ground-truth box parameters:
\begin{equation}
L_{bbox} = \frac{1}{N_{pos}} 
\sum_{(x,y)\in Pos} 
\left\| \mathrm{BBox}^{gt}_{xy} - \mathrm{BBox}^{pred}_{xy} \right\|_1
\end{equation}
where $\|\cdot\|_1$ denotes the L1 norm, and $\mathrm{BBox}^{gt}_{xy}$ and $\mathrm{BBox}^{pred}_{xy}$ represent the ground-truth and predicted bounding boxes at position $(x,y)$, respectively. 
This loss function enables the detection head to more accurately regress the center point offset and the width or height information of the target area, working in tandem with the heatmap target to achieve complete regional localization.

\paragraph{GMoE Gating Entropy Regularization}  
To prevent gated networks from prematurely favoring certain expert paths during training, which leads to rigid routing choices (routing collapse) \cite{ref47}, \cite{ref50}, their information entropy can be defined as:
\begin{equation}
H\left(g(x_{r,l,g})\right) = -\sum_{i=1}^{k} g_i(x_{r,l,g}) \log g_i(x_{r,l,g})
\end{equation}
For all inputs $x_{r,l,g}$, compute the average gated entropy and add the weight coefficient to form the final regularization loss term:
\begin{equation}
\begin{split}
L_{\text{entropy}}
&= -\lambda \cdot \frac{1}{G} \sum_{g=1}^{G} H\!\left(g(x_{r,l,g})\right) \\
&= \lambda \cdot \frac{1}{G} \sum_{g=1}^{G} \sum_{i=1}^{k} g_i(x_{r,l,g}) \log g_i(x_{r,l,g})
\end{split}
\end{equation}
where $\lambda$ denotes the entropy regularization coefficient. This loss term effectively promotes the gated network to utilize different expert resources more rationally during the early training stages, maintaining diversity and balance in expert usage. This enhances the differentiation of expert functions, thereby improving the model's adaptability and generalization capabilities for complex image regions.
Through the design and implementation of the aforementioned optimization strategy, the model achieves more effective object-level cross-view region localization while maintaining training stability, thereby attaining higher accuracy and generalization capabilities.

\section{Experiments}\label{sec:experiments}
\subsection{Experimental Settings}

\paragraph{Implementation Details}
To systematically validate the effectiveness of the proposed method in cross-view object localization tasks, evaluation experiments were conducted on the CVOGL\_DroneAerial dataset. The dataset was split into training, validation, and test sets at a 7:2:1 ratio, ensuring consistent category distribution across subsets. All experiments were conducted on an NVIDIA RTX 4090 GPU platform using the pytorch framework. The adam optimizer was employed with an initial learning rate of $1\times10^{-4}$, batch size of 8, and 25 training epochs. A "warm-up + cosine decay” strategy was employed for learning rate scheduling. To mitigate overfitting, multiple data augmentation techniques—including random scaling, image flipping, and cropping—were integrated during training. These were combined with the DropPath mechanism to enhance model robustness. The backbone network is based on the swin transformer architecture, featuring a four-stage hierarchical design and incorporating a GMoE to boost feature selectivity and expressiveness. Input consists of drone image and satellite image pairs, both uniformly resized to $1024\times1024$ resolution. A $4\times4$ patch embedding with a 96-dimensional embedding vector serves as the encoding starting point, enabling precise modeling of spatial structural differences between cross-view images.

\paragraph{Evaluation Settings}
To comprehensively evaluate the model's performance in cross-view object localization tasks, this paper employs three complementary evaluation metrics: mean intersection over union (mIoU), accuracy@0.25, and accuracy@0.5 (denoted as acc@0.25 and acc@0.5). These metrics are widely adopted in cross-view geo-localization tasks, effectively reflecting a model's spatial alignment capability with the target area and the accuracy of regional boundary localization.

The mIoU serves as a continuous measure of the overlap between prediction and ground truth regions, defined as:
\begin{equation}
\text{IoU}_i = \frac{|P_i \cap G_i|}{|P_i \cup G_i|}, 
\qquad
\text{mIoU} = \frac{1}{N} \sum_{i=1}^{N} \text{IoU}_i
\end{equation}
where $P_i$ and $G_i$ denote the predicted and ground-truth target regions of the $i$-th image, respectively; $|\cdot|$ indicates the pixel area; and $N$ is the total number of test images.

To further evaluate the model's localization capability under different tolerance levels, this paper introduces two IoU-based discrimination metrics: acc@0.25 and acc@0.5. These metrics define “correct localization” when the IoU between the predicted region and the ground truth region exceeds 0.25 and 0.5, respectively. Their definitions are as follows:
\begin{equation}
\text{acc@t} = \frac{1}{N} \sum_{i=1}^{N} {1}(\text{IoU}_i > t), 
\qquad t \in \{0.25, 0.5\}
\end{equation}
where $\mathbf{1}(\cdot)$ is the indicator function, returning 1 if the condition holds and 0 otherwise. Specifically, acc@0.25 evaluates the model's detection capability under a more tolerant error margin, making it suitable for coarse matching performance assessment, particularly when the targets are large in size or significant viewpoint variations exist. In contrast, acc@0.5 emphasizes precise boundary localization, reflecting strict localization capability. These two metrics, when used together, not only reflect the model's performance across both coarse and fine-grained localization tasks but also facilitate cross-comparison of model effectiveness across different architectures and design strategies.

\begin{table}[!t]
\centering
\caption{Overall performance (\%) of different methods on CVOGL\_DroneAerial.}
\label{tab:overall_cvogldrone}
\renewcommand{\arraystretch}{1.15}
\setlength{\tabcolsep}{3.6pt}
\begin{tabular*}{0.98\columnwidth}{@{\extracolsep{\fill}}lcccccc@{}}
\toprule
\multirow{2}{*}{\raisebox{-2.0\height}{Method}} &
\multicolumn{3}{c}{test} &
\multicolumn{3}{c}{validation} \\
\cmidrule(lr){2-4}\cmidrule(lr){5-7}
 & \makecell[c]{acc@0.25\\(\%)} & \makecell[c]{acc@0.5\\(\%)} & \makecell[c]{mIoU\\(\%)} &
   \makecell[c]{acc@0.25\\(\%)} & \makecell[c]{acc@0.5\\(\%)} & \makecell[c]{mIoU\\(\%)} \\
\midrule
SAFA      & 35.25 & 5.68 & 17.53 & 32.36 & 4.31 & 16.28 \\
DSTG      & 36.98 & 5.68 & 18.25 & 33.83 & 5.22 & 17.37 \\
TransGeo  & 37.25 & 5.42 & 16.03 & 35.69 & 4.62 & 16.15 \\
DINO-MSRA & 22.28 & 3.21 & 11.02 & 20.25 & 4.36 & 14.53 \\
CAMP      & 20.21 & 3.21 & 13.88 & 18.53 & 2.58 & 13.68 \\
DetGeo    & 61.97 & 5.81 & 21.50 & 59.81 & 5.15 & 12.14 \\
\textcolor{paperpurple}{\textbf{SMGeo}} &
\textcolor{paperpurple}{\textbf{87.51}} &
\textcolor{paperpurple}{\textbf{62.50}} &
\textcolor{paperpurple}{\textbf{61.45}} &
\textcolor{paperpurple}{\textbf{85.25}} &
\textcolor{paperpurple}{\textbf{58.96}} &
\textcolor{paperpurple}{\textbf{59.36}} \\
\bottomrule
\end{tabular*}
\end{table}
\subsection{Overall Performance Evaluation and Analysis}
To comprehensively and objectively evaluate the effectiveness of the proposed method (SMGeo) in object-level cross-view localization tasks, we conducted rigorous comparisons with several representative methods under identical experimental settings using the CVOGL\_DroneAerial dataset. The comparison methods include image retrieval-based approaches such as SAFA \cite{ref51}, DSTG \cite{ref52}, TransGeo \cite{ref15}, NINO-MSRA \cite{ref26}, CAMP \cite{ref53}, and the latest detection-based method DetGeo \cite{ref20}. Since most existing cross-view research methods focus on whole-image retrieval tasks rather than object-level regional localization, directly comparing localization accuracy presents challenges. To ensure fair comparison across different paradigms, we employed the transformation scheme referenced in \cite{ref33}, implementing the following processing steps to enable reasonable and rigorous evaluation of all methods at the same scale and under identical evaluation metrics.

First, for each query image, we use image retrieval methods to obtain the location of the most matching satellite image region. Then, we use the geometric center point of this matched satellite image region as the predicted center position for target localization. Next, based on the width and height of the ground truth bounding box in the query image, we generate a predicted bounding box with this center point as its geometric center. Finally, the resulting predicted bounding box is compared with the ground truth bounding box to compute the mean overlap ratio (mIoU) and accuracy metrics under different tolerances (acc@0.25, acc@0.5). Through this process, methods originally designed for image-level matching are uniformly mapped to the object-level localization task, establishing a preliminary, comparable experimental framework \cite{ref53}.

\begin{figure*}[!t]  
\centering
\includegraphics[width=\linewidth, keepaspectratio]{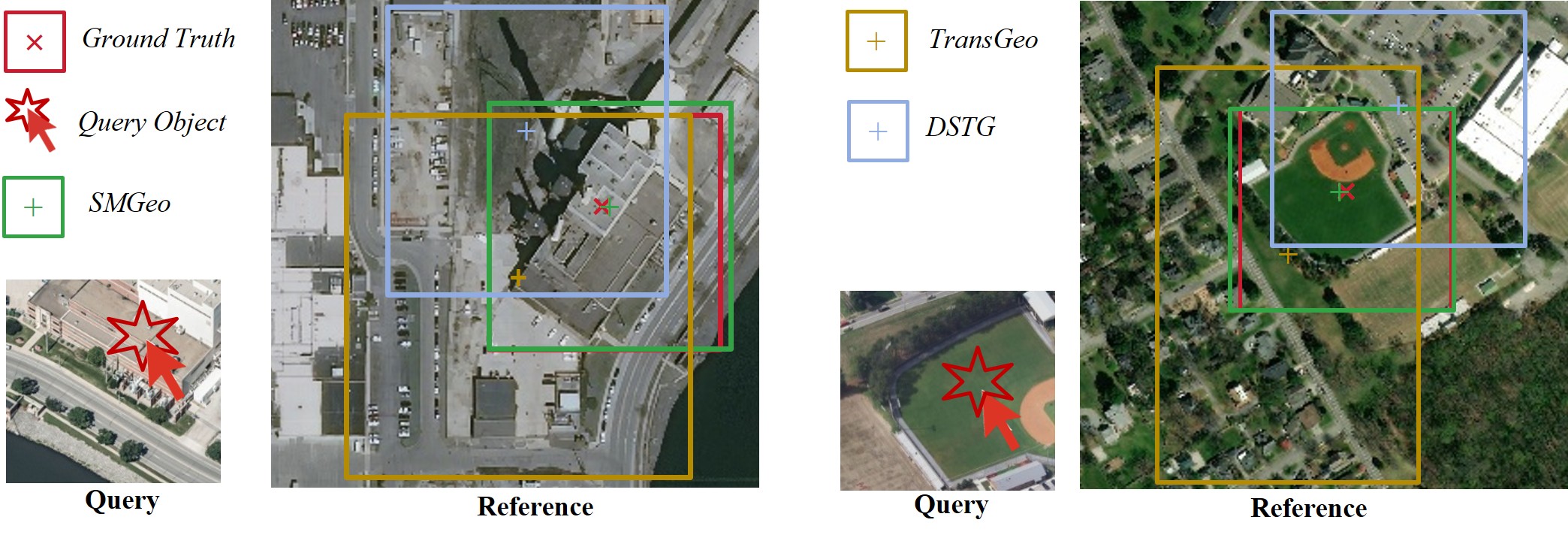}
\caption{Comparison of object geo-localization results across different cross-view methods. The figure displays geolocation results from TransGeo, DSTG, and SMGeo methods. The red bounding box indicates the true object location.}
\label{fig_8}
\end{figure*}
\begin{figure*}[!t]  
\centering
\includegraphics[width=\linewidth, keepaspectratio]{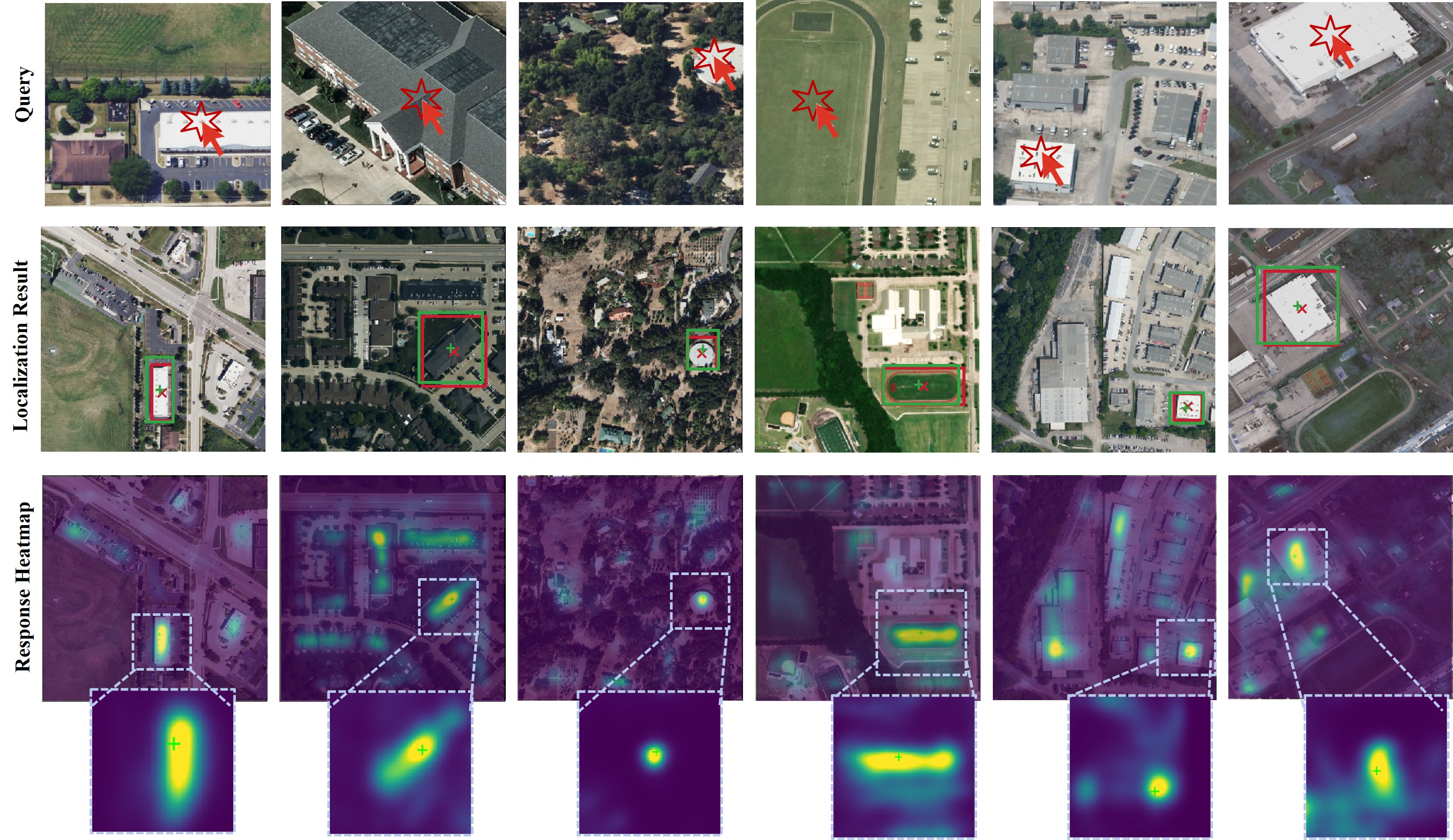}
\caption{Heatmap visualization and localization result example. Each column shows the query image, reference image, response heatmap on the reference, and the final localization result image in sequence.}
\label{fig_9}
\end{figure*}

The detailed performance comparison of all methods obtained in our experiments is shown in Table~\ref{tab:overall_cvogldrone}. From the results, it is evident that on the test split, the proposed SMGeo achieves an accuracy of 87.51\% on acc@0.25, surpassing the closest competitor DetGeo (61.97\%) by 25.54 percentage points. This fully illustrates that the SMGeo network can more accurately capture coarse-scale localization information of targets. Under the stricter acc@0.5 evaluation criterion, the SMGeo method still demonstrates an absolute advantage, achieving an accuracy of 62.50\%, far exceeding the second-place DetGeo method at 57.66\%, and significantly outperforming other mainstream methods such as SAFA (5.68\%), DSTG (6.58\%), and TransGeo (5.42\%). This result validates the powerful adaptability of the proposed dynamic feature fusion mechanism and GMoE network architecture in high-precision localization scenarios. Furthermore, regarding the mIoU metric, the SMGeo method significantly outperformed others with a score of 61.45\%, where the second-best DetGeo method achieved only 54.05\%. This outcome demonstrates that the SMGeo model not only accurately determines the approximate location of targets but also precisely predicts their exact boundaries, reflecting its robust spatial feature learning and fusion capabilities. Furthermore, we conducted a comprehensive evaluation on the validation set to further validate the method's generalization performance and robustness. SMGeo's performance metrics on the validation set also demonstrated a clear lead. Where acc@0.25 reached 85.25\%, while the closest competitor, DetGeo, achieved 59.81\%. Under the stringent acc@0.5 metric, SMGeo reached 58.96\%, still showing a significant improvement over DetGeo (55.15\%). Regarding the mIoU metric, SMGeo achieved 59.36\% on the validation set, substantially exceeding DetGeo's 52.14\% and all other comparison methods. These results demonstrate that the proposed SMGeo method exhibits an absolute leading advantage in cross-view object localization tasks.

Although the performance evaluation metrics in this paper have clearly demonstrated SMGeo's superiority, the intrinsic reasons behind its fundamental advantages warrant further analysis. We observe that existing methods such as SAFA, DSTG, TransGeo, DINO-MSRA, and CAMP are primarily designed for image-level matching based on global visual features, lacking the capability to precisely capture fine-grained features at the object level. In contrast, the proposed SMGeo specifically employs a unified swin transformer architecture combined with GMoE for object-level feature modeling—a key factor behind its substantial performance lead.

Specifically, as shown in Fig.~\ref{fig_8}, SMGeo can accurately predict the accuracy location of the target.  Its predicted bounding box (green dashed line) closely matches the true bounding box (red solid line), and the predicted centers point (green ``+'') maintains minimal deviation from the true center point (red ``$\times$''), intuitively showcasing SMGeo’s precision at the object level. This accuracy stems primarily from the anchor-free detector head designed in this paper, which directly learns the mapping relationship between target center positions and target scales, rather than relying on a simple center point estimation strategy based on whole-image matching.

Furthermore, as analyzed in Fig.~\ref{fig_9}, SMGeo effectively focuses on object regions rather than the global features of the image during cross-view tasks. The prominent bright areas in the heatmaps (with regional peaks marked by green ``+'') accurately cover the target detection regions. This demonstrates that the proposed dynamic feature fusion and GMoE architecture adaptively enhance attention to object regions, enabling the network to capture local spatial information more accurately~\cite{ref54}. This object-level feature modeling capability is exactly what other image-retrieval methods lack and is a fundamental reason why SMGeo performs exceptionally well in cross-view object localization.

\begin{figure}[t] 
\centering
\includegraphics[width=\linewidth]{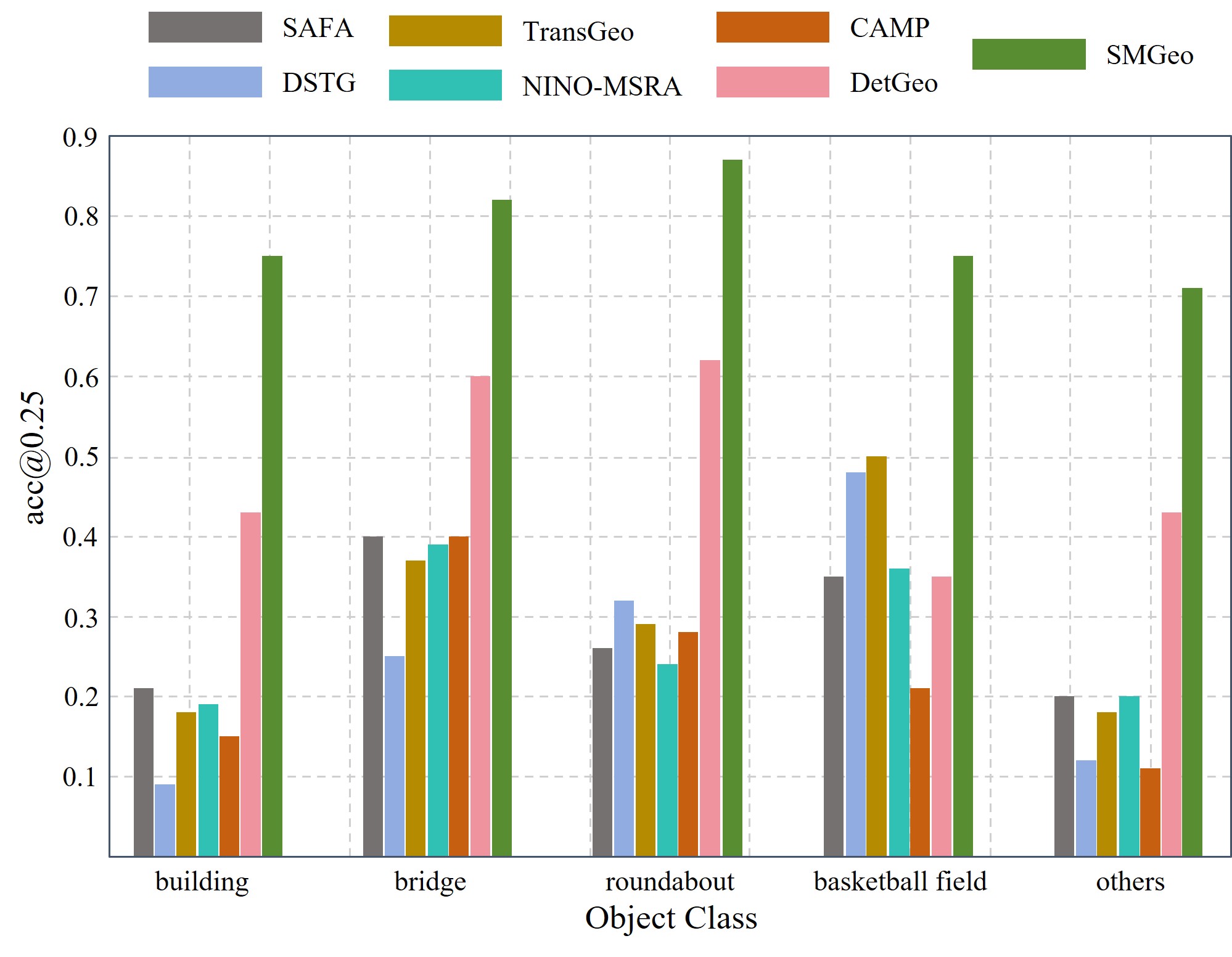}
\caption{Performance (accuracy\%) of different object classes in cross-view tasks.}
\label{fig_10}
\end{figure}
\begin{figure}[t] 
\centering
\includegraphics[width=\linewidth]{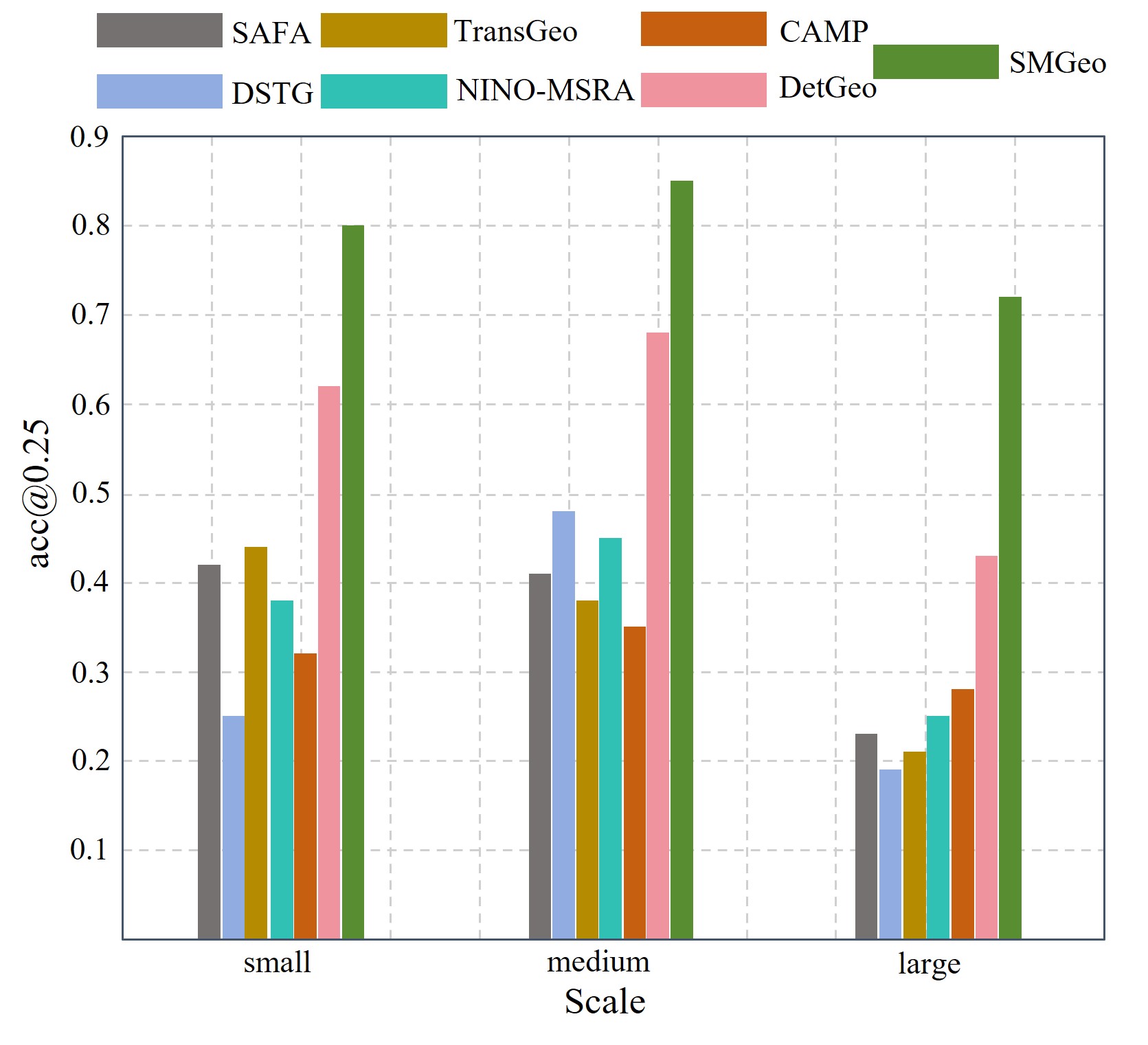}
\caption{Performance (accuracy\%) of different sizes of regions in cross-view tasks.}
\label{fig_11}
\end{figure}
In summary, SMGeo's performance advantages extend beyond surface metrics. More significantly, its unified shared network architecture, object-level feature modeling approach, and dynamic feature fusion mechanism effectively address the challenge of insufficient accuracy in learning precise object features during cross-view object localization tasks. This enables the model not only to accurately pinpoint target regions but also lays a theoretical foundation for more complex remote sensing and unmanned system perception applications.

\subsection{Category and Region Scale Analysis}
To further validate the effectiveness and robustness of the proposed method for object-level localization tasks across different categories and region sizes, we conducted more detailed comparative experiments on various classification objects and regions. Specifically, we perform in-depth experiments and analysis from two dimensions: “object category” and “region scale”.

We selected representative target categories that are typical and common in drone imagery, including buildings, roundabouts, basketball courts, bridges, and other objects. We evaluated the localization performance of SMGeo against other advanced methods for each category to analyze performance differences across target types. In addition, we categorized image object regions by size into three scales: small-scale (smaller than $300 \times 300$), medium-scale (between $300 \times 300$ and $512 \times 512$), and large-scale (larger than $512 \times 512$). This allowed us to thoroughly examine the models' adaptability and robustness toward target regions of varying scales.

\begin{figure*}[!t]  
\centering
\includegraphics[width=\linewidth, keepaspectratio]{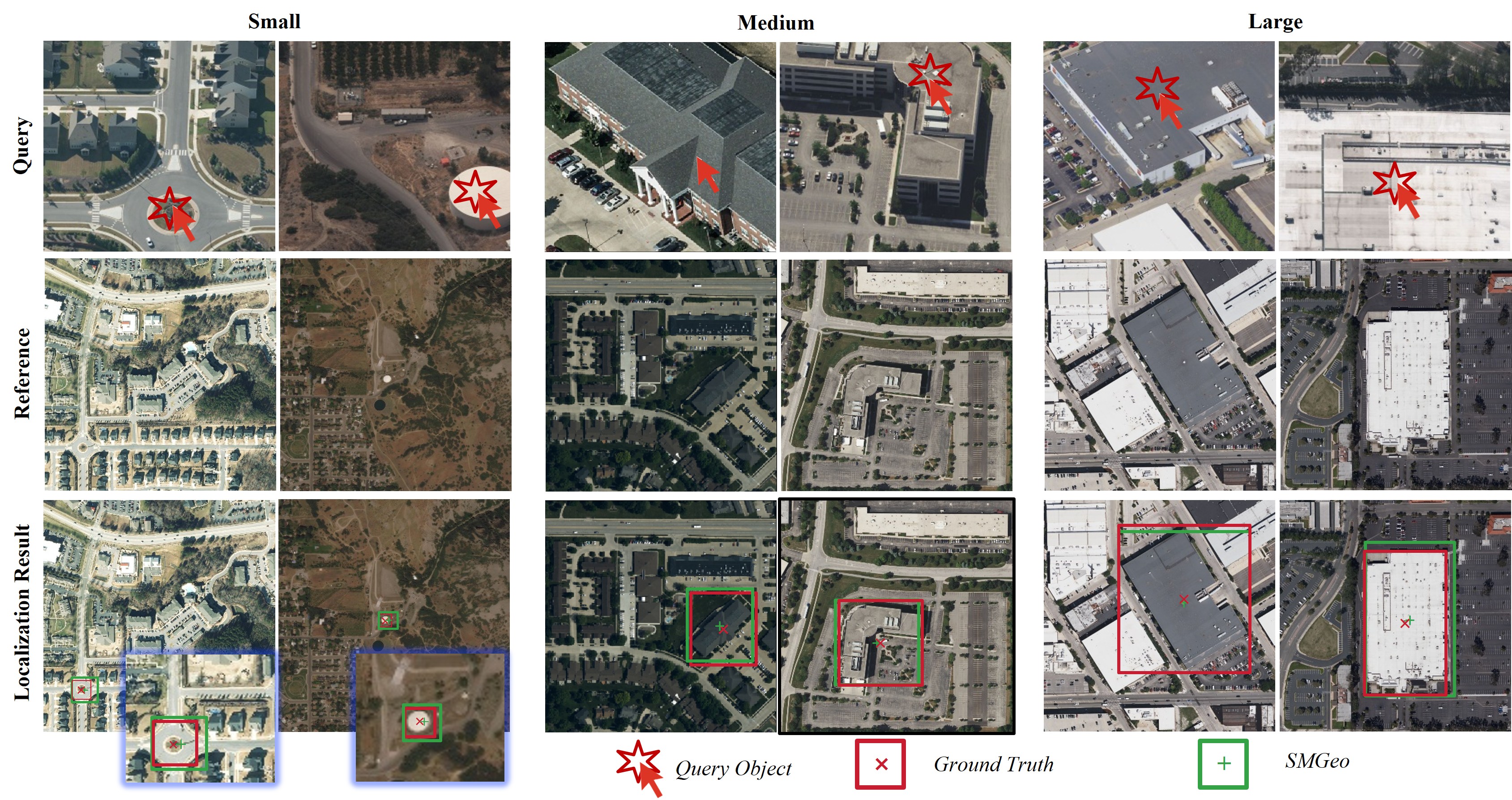}
\caption{Output results of the SMGeo method for tasks at different sizes of regions. Each column represents small, medium, and large scales respectively; from top to bottom, they correspond to the query image, satellite reference image, and positioning results.}
\label{fig_12}
\end{figure*}
We employ acc@0.25 as the primary evaluation metric for this experiment, and all results are reported on the test set. We visualize the results along the above two dimensions using bar charts, forming a unified comparison (as illustrated in Fig.~\ref{fig_10} and Fig.~\ref{fig_11}). This diagram intuitively presents the localization accuracy differences among SMGeo and methods such as SAFA, DSTG, TransGeo, DINO-MSRA, CAMP, and DetGeo under different categories and region scales. The results demonstrate that in the building category, the SMGeo method achieves significantly higher average accuracy than mainstream approaches such as DetGeo and TransGeo. In the more complex categories of stadiums and bridges, SMGeo’s lead further expands, indicating its remarkable adaptability to intricate scenes and detailed objects. In particular, for the basketball field category, SMGeo method outperforms the second-best method (DetGeo) by approximately 30 percentage points on acc@0.25. This demonstrates that the proposed dynamic feature fusion mechanism and GMoE design exhibit distinct advantages in identifying spatial details within complex target regions. For regions of varying scales (as shown in Fig.~\ref{fig_12}), small regions often possess fewer feature information and are difficult to localize, while large regions are prone to containing confusing information, posing challenges for both models. Experimental results demonstrate that the SMGeo method achieves the highest accuracy metrics across all three scale categories. Under the most challenging small-area conditions, the SMGeo method outperforms the second-best method, DetGeo, by approximately 18 percentage points. It also exhibits more stable performance under large-area conditions. These results indicate that the single-branch GMoE architecture proposed by SMGeo can effectively adapt to spatial variations across scales, enabling more robust and precise localization.

\subsection{Ablation Study and Analysis}

To thoroughly investigate the specific impact of the GMoE designed in the SMGeo method on improving performance for cross-view object localization tasks, this section conducts detailed ablation experiments (all results obtained on the test set) centered on configurations related to the GMoE mechanism. We systematically discuss the findings by integrating visualizations of grid-level expert activations with a mechanistic analysis. Additionally, we provide visual analyses of grid-level expert activations to more intuitively reveal the intrinsic working mechanisms and division of labor characteristics within the expert system.

\begin{table}[!t]
\centering
\caption{Effect of the number of experts on performance and complexity.}
\label{tab:num_experts}
\renewcommand{\arraystretch}{1.15}
\setlength{\tabcolsep}{3.6pt}
\begin{tabular*}{0.98\columnwidth}{@{\extracolsep{\fill}}cccccc@{}}
\toprule
\makecell[c]{number\\of experts} &
\makecell[c]{acc@0.25\\(\%)} &
\makecell[c]{acc@0.5\\(\%)} &
\makecell[c]{mIoU\\(\%)} &
FLOPs & \#P \\
\midrule
2 & 83.51 & 57.23 & 57.71 & 318G    & 30.61M \\
3 & 84.23 & 57.92 & 58.42 & 319G    & 31.35M \\
4 & 84.76 & 58.44 & 58.98 & 319G    & 32.09M \\
5 & 85.05 & 58.71 & 59.26 & 320G    & 32.83M \\
\textcolor{paperpurple}{\textbf{6}} &
\textcolor{paperpurple}{\textbf{85.25}} &
\textcolor{paperpurple}{\textbf{58.96}} &
\textcolor{paperpurple}{\textbf{59.34}} &
\textcolor{paperpurple}{\textbf{320.46G}} &
\textcolor{paperpurple}{\textbf{33.57M}} \\
7 & 85.26 & 58.97 & 59.34 & 321G    & 34.31M \\
8 & 84.26 & 58.97 & 59.34 & 321.53G & 35.05M \\
\bottomrule
\end{tabular*}
\end{table}

\begin{table}[!t]
\centering
\caption{Effect of the top-k experts on performance and complexity.}
\label{tab:topk_experts}
\renewcommand{\arraystretch}{1.15}
\setlength{\tabcolsep}{3.6pt}
\begin{tabular*}{0.98\columnwidth}{@{\extracolsep{\fill}}cccccc@{}}
\toprule
\makecell[c]{top-\emph{k}\\experts} &
\makecell[c]{acc@0.25\\(\%)} &
\makecell[c]{acc@0.5\\(\%)} &
\makecell[c]{mIoU\\(\%)} &
\makecell[c]{FLOPs\\\phantom{(\%)}} &
\makecell[c]{\#P\\\phantom{(\%)}} \\
\midrule
1 & 84.71 & 58.41 & 58.86 & 318G    & 33.57M \\
\textcolor{paperpurple}{\textbf{2}} &
\textcolor{paperpurple}{\textbf{85.25}} &
\textcolor{paperpurple}{\textbf{58.96}} &
\textcolor{paperpurple}{\textbf{59.36}} &
\textcolor{paperpurple}{\textbf{320.52G}} &
\textcolor{paperpurple}{\textbf{33.57M}} \\
3 & 85.25 & 58.96 & 59.36 & 323G    & 33.57M \\
4 & 85.26 & 58.96 & 59.36 & 326G    & 33.57M \\
5 & 85.26 & 58.96 & 59.36 & 329G    & 33.57M \\
\bottomrule
\end{tabular*}
\end{table}
\paragraph{Analysis of Number of Experts}
In the grid-level MoE module of SMGeo, the number of experts directly determines the number of feature representations that can be scheduled for each spatial grid. Different numbers of experts also exert varying effects on model parameter scale and performance \cite{ref46}. To address this, we analyzed the specific impact of varying expert counts on cross-view object localization performance and model parameter scaling. To evaluate these effects, we set expert counts to $\{2,3,4,5,6,7,8\}$, selecting the top-2 experts for computation at each setting. Experimental results are presented in Table~\ref{tab:num_experts}.

\begin{figure*}[!t]  
\centering
\includegraphics[width=\linewidth]{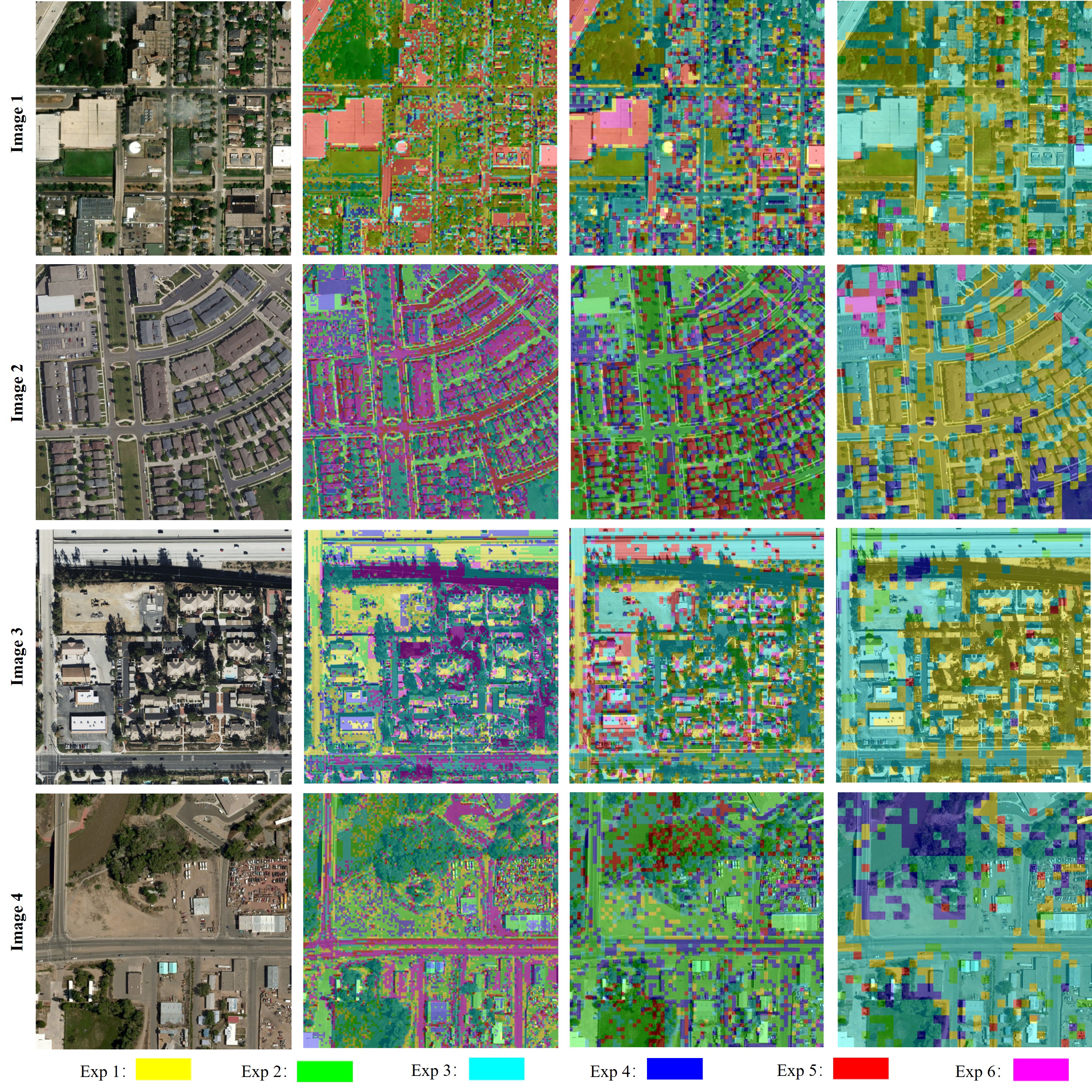}
\caption{Grid-level expert activation visualization. The left panel shows the reference satellite imagery; the right panel displays three columns representing the grid-level expert activation distributions for Stage 1, Stage 2, and Stage 3, respectively. Each grid cell is colored according to its assigned primary expert (color scale 1–6), with color serving as a category indicator.}
\label{fig_13}
\end{figure*}

From Table~\ref{tab:num_experts}, when the number of experts increases from 2 to 6, the performance metrics improve significantly. This indicates that increasing the number of experts effectively enhances the model's ability to represent complex spatial features. When the number of experts further increases to 8, although the model's performance improves, the rate of improvement tends toward saturation. Simultaneously, the number of model parameters increases significantly, raising computational overhead. This indicates that simply increasing the number of experts does not lead to unlimited performance gains; a trade-off between performance benefits and computational cost must be considered. Based on the above analysis, we ultimately determine that 6 experts is the optimal choice, as this setting achieves the best balance between model performance and parameter overhead.

\begin{table*}[!t]
\centering
\begin{threeparttable}
\caption{Ablation on GMoE insertion across stages (\%).}
\label{tab:moe_stage_insertion}
\renewcommand{\arraystretch}{1.25}
\setlength{\tabcolsep}{5pt} 
\begin{tabular*}{0.92\textwidth}{@{\extracolsep{\fill}}ccccccccc@{}} 
\toprule
Stage0 & Stage1 & Stage2 & Stage3 & acc@0.25(\%) & acc@0.5(\%) & mIoU(\%) & FLOPs & \#P \\
\midrule
None & None & None & None & 80.01 & 55.60 & 54.02 & 315.02G & 29.13M \\
None & None & None & Even & 83.50 & 57.53 & 57.81 & 316.37G & 30.24M \\
None & None & Even & Even & 85.06 & 58.80 & 59.25 & 319.26G & 32.15M \\
\textcolor{paperpurple}{\textbf{None}} &
\textcolor{paperpurple}{\textbf{Even}} &
\textcolor{paperpurple}{\textbf{Even}} &
\textcolor{paperpurple}{\textbf{Even}} &
\textcolor{paperpurple}{\textbf{85.25}} &
\textcolor{paperpurple}{\textbf{58.96}} &
\textcolor{paperpurple}{\textbf{59.36}} &
\textcolor{paperpurple}{\textbf{320.46G}} &
\textcolor{paperpurple}{\textbf{33.57M}} \\
Even & Even & Even & Even & 84.53 & 58.51 & 58.87 & 323.19G & 35.79M \\
All & All & All & All & 83.04 & 57.02 & 57.51 & 331.38G & 42.45M \\
\bottomrule
\end{tabular*}

\begin{tablenotes}[flushleft]
\footnotesize
\item \textit{Notes}: None\textemdash{}no insertion; Even\textemdash{}insert only in even-indexed blocks of the stage; All\textemdash{}insert in all blocks of the stage. Best values are in purple. The best setting is Stage0: None, Stages 1--3: Even.
\end{tablenotes}
\end{threeparttable}
\end{table*}
\paragraph{Analysis of top-\texorpdfstring{$k$}{k} Experts Selection}
Another key adjustable parameter in the GMoE mechanism is the top-k expert count, which refers to the actual number of experts utilized during each forward propagation. the impact of this parameter on model localization performance in detail, we tested performance metrics under different settings from top-1 to top-5 while keeping the total number of experts fixed at 6. The experimental results are shown in Table~\ref{tab:topk_experts}.

As shown in Table~\ref{tab:topk_experts}, when only one expert is selected, the model performance is significantly inadequate, indicating that a single expert has limited processing capacity and cannot fully express complex target features. When the number of selected experts increases to 2, all performance metrics of the model show a significant improvement, demonstrating that the multi-expert collaboration mechanism effectively enhances the model's ability to express cross-view image features \cite{ref45}. Although theoretically providing richer expert information, further increasing to three or four experts yielded only marginal performance gains in practical experiments. Moreover, the increase in FLOPs significantly prolonged inference time. Therefore, this paper ultimately determines the top-2 experts selection as the optimal configuration, balancing both localization accuracy and inference efficiency.

\paragraph{Grid-Level Expert Activation Visualization}
To further investigate the impact of GMoE at different network levels on the overall performance and parameter scale of SMGeo, we also designed systematic ablation experiments. We insert GMoE modules into various stages of the swin transformer, with results shown in Table~\ref{tab:moe_stage_insertion}.

From the Table~\ref{tab:moe_stage_insertion}, we can observe that the GMoE mechanism is most suitable for insertion at mid-to-high layers (Stage 1 to Stage 3). At these stages, after multiple experts have achieved a certain level of spatial and semantic abstraction, dynamic expert division maximizes the model's adaptability to complex regions and diverse semantics. Conversely, inserting GMoE at shallow layers, where expert division has not yet been established, limits the network's ability to model basic textures and edges, thereby increasing parameter and computational overhead. Simultaneously, excessive GMoE stacking or shallow-to-deep global insertion leads to “routing dilution” and diminished generalization capabilities\cite{ref55,ref56}. The optimal configuration ensures sufficient regional semantic information at mid-to-high levels, enabling thorough GMoE division and routing while balancing localization accuracy and resource efficiency.

To further illustrate the specific division of labor and roles of GMoE module in cross-view object localization tasks, we conducted a visualization analysis of grid-level expert activations. The expert activations across the final three stages of the SMGeo model are displayed, forming a grid-level expert activation visualization as shown in Fig.~13.

Through the visualization analysis in Fig.~\ref{fig_13}, we observe that SMGeo exhibits distinct spatial specialization across different stages, with each expert tending to process specific local patterns or semantic features. For instance, Expert~1 primarily handles to building edges, Expert~3 focuses on small-scale structures, while Expert~6 excels in complex textured regions. The division of labor becomes more refined in later stages, particularly in Stage~3, where the activation regions of each expert exhibit distinct semantic distinctions. This demonstrates the adaptive fusion mechanism's capability for precise spatial representation of cross-view image pair. This clear division of labor intuitively validates the effectiveness and rationality of the multi-expert hybrid mechanism for cross-view object localization, further elucidating the underlying mechanism behind the performance improvements achieved by the proposed method.

In summary, the ablation experiments and analysis results in this section fully demonstrate the critical influence of key parameters in the proposed SMGeo method on model performance. Furthermore, expert activation visualization provides an in-depth analysis of the model's internal mechanisms. Furthermore, this visualization analysis offers more intuitive guidance for subsequent model optimization, aiding in further enhancing the performance of GMoE in complex multi-view localization tasks.

\begin{figure}[t]
\centering
\includegraphics[width=\linewidth]{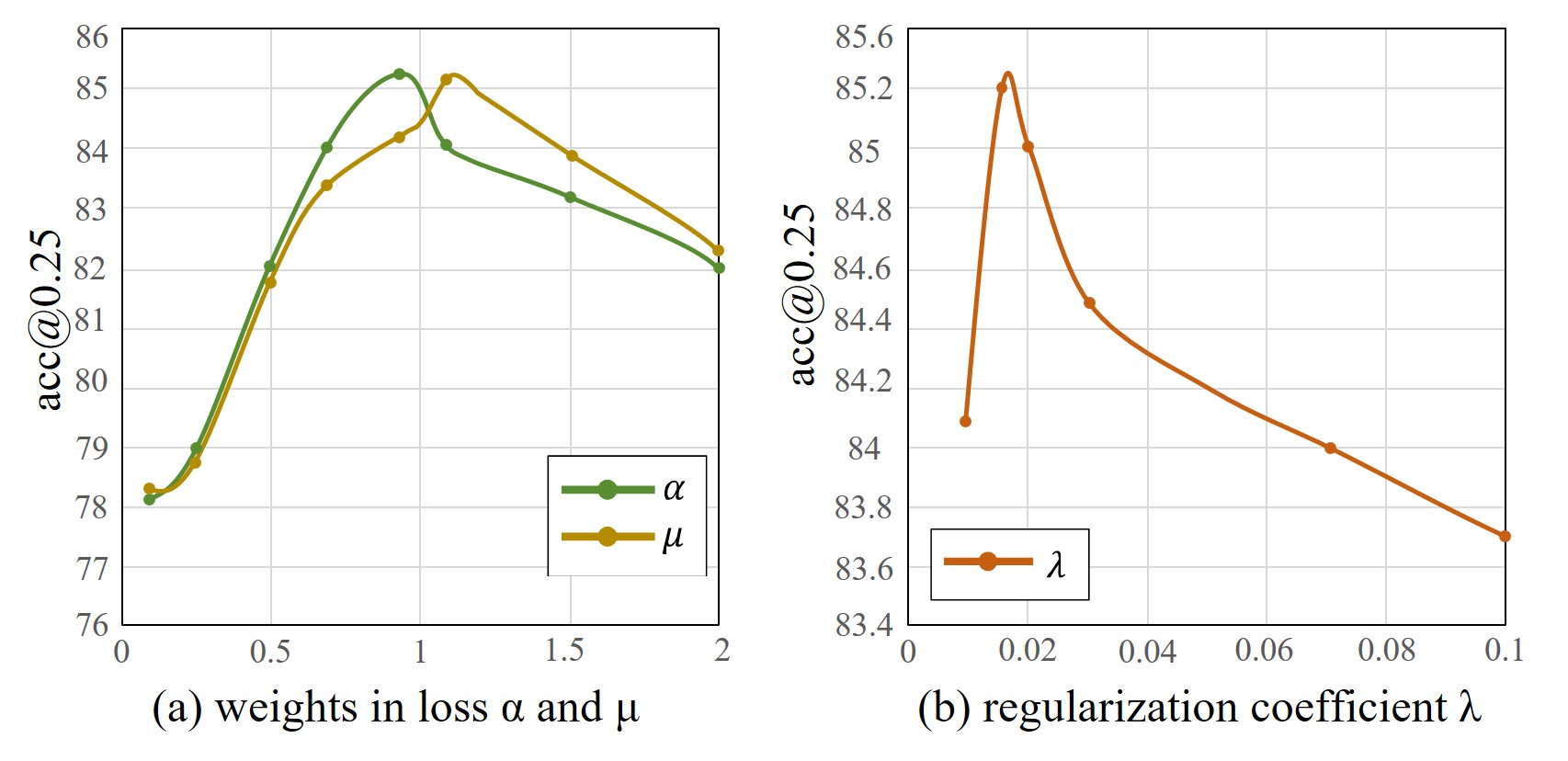}
\caption[Hyper-parameter sensitivity (acc@0.25)]%
{Hyper-parameter sensitivity (acc@0.25). The two plots show the effect of $\alpha$, $\mu$, and $\lambda$ on performance; the best results occur at $\alpha=0.9$, $\mu=1.1$, and $\lambda=0.015$.}
\label{fig_14}
\end{figure}

\subsection{Parameter Analysis}
To systematically analyze the impact of key hyperparameters on the final performance of SMGeo, we conducted a systematic analysis of three core hyperparameters: the center-heatmap loss weight $\alpha$, the bounding-box regression loss weight $\mu$, and the gating-entropy regularization coefficient $\lambda$. These hyperparameters directly influence  the model's training stability and localization accuracy. While keeping other hyperparameters fixed, we varied $\alpha$ in the range $[0.1, 2.0]$, $\mu$ in the range $[0.1, 2.0]$, and $\lambda$ in the range $[0.001, 0.1]$. The corresponding experimental results are shown in Fig.~\ref{fig_14}.

The hyperparameter $\alpha$ balances the loss of the confidence branch in the detection head, determining the model's judgment of “object existence” in the anchor-free architecture. The hyperparameter $\mu$ controls the spatial localization accuracy of the predicted bounding boxes, and $\lambda$ encourages distribution diversity in GMoE routing, suppressing premature convergence and promoting balanced utilization of expert networks. Specifically, the model achieves its best performance when $\alpha = 0.9$. Too low a value leads to inaccurate center point predictions, compromising overall localization accuracy; too high a value may cause excessive focus on center points, impairing the model's ability to distinguish key features. When $\mu = 1.1$, the model achieves optimal performance. Too low a value risks neglecting positional accuracy, while too high a value may suppress learning in other branches, leading to unstable training. When $\lambda = 0.015$, localization accuracy improves; excessive regularization suppresses the model's exploration capability, resulting in performance degradation. Ultimately, we determined that the model achieves optimal performance when $\alpha = 0.9$, $\mu = 1.1$, and $\lambda = 0.015$.

\section{Conclusion}\label{sec:conclusion}
This paper proposes the SMGeo unified framework for cross-view object localization in drone and satellite applications. It employs a swin transformer with shared weights to encode images from two perspectives, achieving feature alignment within a unified representation space. It introduces GMoE for region-adaptive modeling, combined with query-guided cross-perspective fusion to suppress background interference. Through anchor-free detection heads, it directly regresses target center and scale, enabling end-to-end cross-perspective object geo-localization prediction. On public datasets, SMGeo achieves state-of-the-art performance across multiple metrics while maintaining stability in complex backgrounds and small-object scenarios. Ablation and visualization results demonstrate the framework's reasonable design, enhancing robustness and interpretability while maintaining computational efficiency. Future research will focus on multi-object data and stronger cross-view adaptability to further improve reliability and deployability in real-world tasks.
\bibliographystyle{IEEEtran}   
\bibliography{refs}            

\begingroup
  \setlength{\parskip}{0pt}
  \setlength{\baselineskip}{13.5pt}

  \begin{IEEEbiography}[{\includegraphics[width=1in,height=1.25in,clip,keepaspectratio]{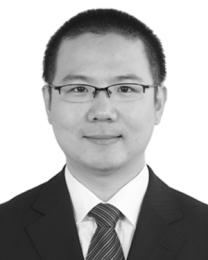}}]{Fan Zhang}
    (Senior Member, IEEE) received the B.E. degree in communication engineering from the Civil Aviation University of China, Tianjin, China, in 2002, the M.S. degree in signal and information processing from Beihang University, Beijing, China, in 2005, and the Ph.D. degree in signal and information processing from the Institute of Electronics, Chinese Academy of Sciences, Beijing, in 2008.

    He is currently a Full Professor of electronic and information engineering with Beijing University of Chemical Technology, Beijing. His research interests are remote sensing image processing, high performance computing, and artificial intelligence.

    Dr. Zhang is an Associate Editor for \textsc{IEEE Access} and a reviewer for \textsc{IEEE Transactions on Geoscience and Remote Sensing}, \textsc{IEEE Journal of Selected Topics in Applied Earth Observations and Remote Sensing}, \textsc{IEEE Geoscience and Remote Sensing Letters}, and \textit{Journal of Radars}.
  \end{IEEEbiography}
  \vspace{-6pt} 

  \begin{IEEEbiography}[{\includegraphics[width=1in,height=1.25in,clip,keepaspectratio]{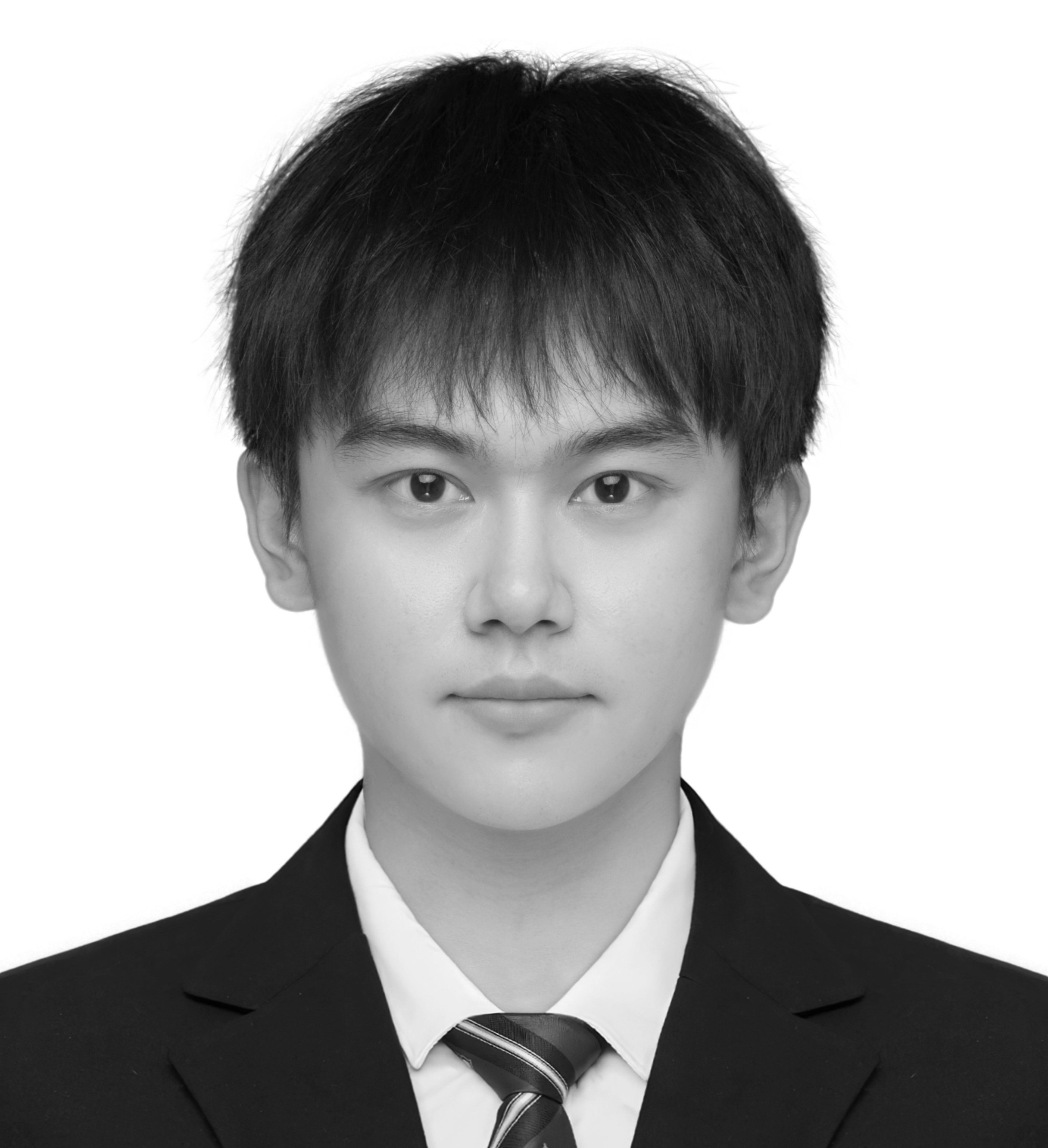}}]{Haoyuan Ren}
    is currently pursuing a master's degree with Beijing University of Chemical Technology, Beijing, China. He is majoring in control science and engineering.

    His research interests include cross-view object geo-localization in remote sensing images and multi-modal object detection.
  \end{IEEEbiography}
  \vspace{200pt}
  \begin{IEEEbiography}[{\includegraphics[width=1in,height=1.25in,clip,keepaspectratio]{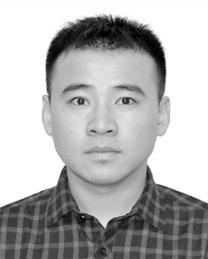}}]{Fei Ma}
    (Member, IEEE) received the B.S., M.S., and Ph.D. degrees in electronic and information engineering from Beihang University, Beijing, China, in 2013, 2016, and 2020, respectively. From 2017 to 2018, he was a Research Fellow with the Department of Electrical Engineering, McGill University, Montreal, QC, Canada.

    He is currently a Full Associate Professor with the College of Information Science and Technology, Beijing University of Chemical Technology, Beijing. His research interests include synthetic aperture radar (SAR) image processing, machine learning, artificial intelligence, and target detection.
  \end{IEEEbiography}
  \vspace{-200pt}

  \begin{IEEEbiography}[{\includegraphics[width=1in,height=1.25in,clip,keepaspectratio]{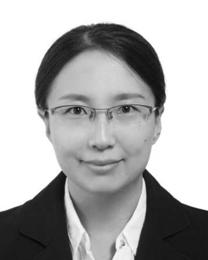}}]{Qiang Yin}
    (Senior Member, IEEE) received the B.S. degrees in electronic and information engineering from Beijing University of Chemical Technology, Beijing, China, in 2004, the M.S. and Ph.D. degrees in signal and information processing from Institute of Electronics, Chinese Academy of Science, Beijing, China, in 2008 and 2016, respectively.

    From 2008 to 2013, she was a research assistant with the Institute of Electronics, Chinese Academy of Sciences, Beijing. From 2014 to 2015, she was a research fellow with European Space Agency, Roma, Italy. Currently, she is an associate professor at the College of Information Science and Technology, Beijing University of Chemical Technology, Beijing, China. Her research interests are polarimetric/polarimetric interferometric SAR and deep learning.
  \end{IEEEbiography}
  \vspace{-200pt}

  \begin{IEEEbiography}[{\includegraphics[width=1in,height=1.25in,clip,keepaspectratio]{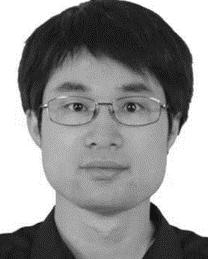}}]{Yongsheng Zhou}
    (Member, IEEE) received the B.E. degree in communication engineering from the Beijing Information Science and Technology University, Beijing, China, in 2005, the Ph.D. degree in signal and information processing from Institute of Electronics, Chinese Academy of Sciences, Beijing, China, in 2010.

    He is currently a Professor of electronic and information engineering at the Beijing University of Chemical Technology, Beijing, China. His general research interests lie in target detection and recognition from microwave remotely sensed image, digital signal and image processing, etc.
  \end{IEEEbiography}

\endgroup

\end{document}